\documentclass{article}

\usepackage[final]{neurips_2021}

\usepackage[utf8]{inputenc} 
\usepackage[T1]{fontenc}    
\usepackage{hyperref}       
\usepackage{url}            
\usepackage{booktabs}       
\usepackage{amsfonts}       
\usepackage{nicefrac}       
\usepackage{microtype}      
\usepackage{xcolor}         
\usepackage{todonotes}
\usepackage[ruled,linesnumbered]{algorithm2e}
\usepackage{siunitx}
\usepackage{amsmath}
\usepackage{amssymb}
\usepackage{amsthm}
\usepackage{stackengine}
\usepackage[caption=false, font=footnotesize]{subfig}

\usepackage{amsmath,amsfonts,bm}

\def\Figref#1{Figure~\ref{#1}}

\def\secref#1{section~\ref{#1}}

\def\eqref#1{equation~\ref{#1}}

\def\algref#1{algorithm~\ref{#1}}

\def\1{\bm{1}}

\DeclareMathAlphabet{\mathsfit}{\encodingdefault}{\sfdefault}{m}{sl}
\SetMathAlphabet{\mathsfit}{bold}{\encodingdefault}{\sfdefault}{bx}{n}

\def\sA{{\mathbb{A}}}

\def\sG{{\mathbb{G}}}

\def\sP{{\mathbb{P}}}

\def\sS{{\mathbb{S}}}

\newcommand{\E}{\mathbb{E}}

\DeclareMathOperator*{\argmax}{argmax}
\DeclareMathOperator*{\argmin}{argmin}

\title{Learning to Execute: Efficiently Learning Universal Plan-Conditioned Policies in Robotics}

\author{Ingmar Schubert$^1$, Danny Driess$^{1}$, Ozgur S. Oguz$^{2}$, and Marc Toussaint$^{1}$ \\
$^1$ Learning and Intelligent Systems Group, TU Berlin, Germany \\
$^2$ Machine Learning and Robotics Lab, University of Stuttgart, Germany \\
\texttt{\{ingmar.schubert@,danny.driess@campus.,toussaint@\}tu-berlin.de}\\
\texttt{ozgur.oguz@ipvs.uni-stuttgart.de}
}

\begin{document}

\maketitle

\begin{abstract}
  Applications of Reinforcement Learning (RL) in robotics are often limited by high data demand.
  On the other hand, approximate models are readily available in many robotics scenarios, making model-based approaches like planning a data-efficient alternative.
  Still, the performance of these methods suffers if the model is imprecise or wrong.
  In this sense, the respective strengths and weaknesses of RL and model-based planners are complementary.
  In the present work, we investigate how both approaches can be integrated into one framework that combines their strengths.
  We introduce Learning to Execute (L2E), which leverages information contained in approximate plans to learn universal policies that are conditioned on plans.
  In our robotic manipulation experiments, L2E exhibits
  increased performance when compared to pure RL, pure planning, or baseline methods combining learning and planning.
\end{abstract}

\section{Introduction}
\label{sec:introduction}
A central goal of robotics research is to design intelligent machines that can solve arbitrary and formerly unseen tasks while interacting with the physical world.
Reinforcement Learning (RL) \citep{sutton2018reinforcement} is a generic framework to automatically learn such intelligent behavior with little human engineering.
Still, teaching an RL agent to actually exhibit general-purpose problem-solving behavior is, while possible in principle, prohibitive in practice.
This is due to practical restrictions including limited computational resources and limited data availability.
The latter limitation is particularly dramatic in robotics, where interaction with the physical world is costly.

On the other hand, for many robotics scenarios, there is a rough model of the environment available.
This can be exploited, e.g., using model-based planning approaches \citep{mordatch2012discovery, kuindersma2016optimization, toussaint2018differentiable}.
Model-based planners potentially offer a more data-efficient way to reason about an agent’s interaction with the world. Model-based planners have been used in many areas of robotics, such as for indoor and aerial robots \citep{faust2018prm}, visual manipulation \citep{jeong2020self}, or humanoid walking \citep{mordatch2015ensemble}.
Still, if the model does not account for stochasticity or contains systematic errors, directly following the resulting plan will not be successful.

The present work starts from the observation that both pure RL approaches and pure planning approaches have strengths and limitations that are fairly complementary.
RL makes no assumptions about the environment but is data-hungry, and model-based planning generally implies model simplifications but is data-efficient.
For robotic manipulation tasks, it seems natural to try and integrate both approaches into one framework that combines the strengths of both.
In the present work we seek to add an additional perspective to the open question of how this can be achieved best.

We introduce a novel approach that we call Learning to Execute (L2E).
Our approach translates sparse-reward \emph{goal-conditioned} Markov Decision Processes (MDPs) \citep{bellman1957markovian} into \emph{plan-conditioned} MDPs.
L2E exploits a simple planning module to create crude plans, which are then used to teach any off-the-shelf off-policy RL agent to execute them.
L2E makes use of final-volume-preserving reward shaping (FV-RS) \citep{schubert2021planbased}, allowing it to train a universal plan-conditioned policy with high data efficiency.
The contributions of this work are:
\begin{itemize}
  \item We introduce L2E, which uses RL to efficiently learn to execute approximate plans from a model-based planner in a plan-conditioned MDP.
  We describe formally how FV-RS can be used as a tool to construct such plan-conditioned MDPs from goal-conditioned MDPs.
  \item We introduce plan replay strategies to efficiently learn universal plan-conditioned policies.
  \item We demonstrate, using robotic pushing problems, that L2E exhibits increased performance when compared to pure RL methods, pure planning methods, or other methods combining learning and planning.
\end{itemize}
We discuss work related to ours in \secref{sec:related_work}, explain background and notation in \secref{sec:background}, and introduce our method in \secref{sec:method_learning_to_execute}.
We present our experimental results in \secref{sec:experiments}, discuss limitations in \secref{sec:discussion}, and conclude with \secref{sec:conclusion}.

\section{Related Work}
\label{sec:related_work}
\subsection{Goal-Conditioned Policies}
Goal-conditioned or universal policies \citep{kaelbling1993learning, moore1999multi, foster2002structure, schaul2015, veeriah2018many, nasiriany2019planning} not only act based on the state the agent finds itself in, but also based on the goal it tries to achieve.
Hindsight Experience Replay (HER) \citep{andrychowicz2017} is a particularly efficient way to learn universal policies.
Here, achieved outcomes of the agent's interaction with the environment are interpreted as desired goals in order to improve sample efficiency in sparse-reward settings.

L2E draws great inspiration from this work, but in contrast to HER, L2E learns a universal plan-conditioned policy.
This means that the L2E policy in general can execute multiple plans leading to the same goal.
Although this presents a more complex learning task, we show in our experiments that by incorporating plan information using plan-based FV-RS, the sample efficiency of L2E is significantly improved over HER.

\subsection{Plan- and Trajectory-Conditioned Policies}
Plan-conditioned policies create behavior that depends on plans that are input to the decision making.
\citet{lynch2020learning} learn plans and how to execute them from data generated by a human ``playing'' with a teleoperated robot.
The resulting policy is conditional on a latent space of encoded plans.
Our work differs from this paradigm in that human interaction is not needed.
Both \citet{lynch2020learning} and \citet{co2018self} directly imitate a planned trajectory by maximizing its likelihood.
In contrast, the plans used in the present work are not directly imitated. Using FV-RS guarantees that the fully trained L2E agent will reach its goal after finite time even if the plan provided is wrong.
\citet{guo2019memory} learn trajectory-conditioned policies to self-imitate diverse (optimal and suboptimal) trajectories from the agent's past experience.

We instead assume in this work that the plan is provided by an external model-based planner.
This allows the L2E agent to use external information during training that could not be concluded from its own experience yet.

\subsection{Learning from Demonstration}
L2E learns how to execute plans in order to achieve different tasks.
In this sense, it is related to Learning from Demonstration (LfD) techniques that exploit demonstrations when learning a task.
Existing work \citep{argall2009survey, hussein2017imitation, ravichandar2020recent} differs significantly both in how the demonstration examples are collected and how the policy is then derived.
\citet{taylor2011integrating} derive an approximate policy from human demonstration, and then use this to bias the  exploration in a final RL stage.
\citet{hester2017deep} train a policy on both expert data and collected data, combining supervised and temporal difference losses.
\citet{salimans2018learning} use a single demonstration as starting points to which the RL agent is reset at the beginning of each episode.
\citet{peng2018deepmimic} use motion capture data to guide exploration by rewarding the RL agent to imitate it.
In \citet{cabi2019scaling}, demonstrations are combined with reward sketching done by a human.
Interactive human feedback during training is another source of information used in  \citet{thomaz2006reinforcement,knox2010combining}.
\citet{kinose2020integration} integrate RL and demonstrations using generative adversarial imitation learning by interpreting the discriminator loss as an additional optimality signal in multi-objective RL.

While these LfD approaches are related to L2E in that external information is used to increase RL efficiency, it is in contrast assumed in L2E that this external information is provided by a planner.

\subsection{Combining Learning with Planning}
Similarly to demonstrations, external plans can be exploited to facilitate learning.
\citet{faust2018prm} connect short-range goal-conditioned navigation policies into complex navigation tasks using probabilistic roadmaps.
In contrast, L2E learns a single plan-conditioned policy for both short-term and long-term decision making.
\citet{sekar2020planning} use planning in a learned model to optimize for expected future novelty. In contrast, L2E encourages the agent to stay close to the planned behavior.
\citet{zhang2016learning} use model-predictive control to generate control policies that are then used to regularize the RL agent.
In L2E, no such intermediate control policy is created, and a reward signal is computed directly from the plan.
In Guided Policy Search \citep{levine2013guided}, differential dynamic programming is used to create informative guiding distributions from a transition model for policy search. These distributions are used to directly regularize the policy in a supervised fashion, while L2E makes use of FV-RS as a mechanism to interface planning and RL.
\citet{christiano2016transfer} learn an inverse dynamics model to transfer knowledge from a policy in the source domain to a policy in the target domain.
The idea of integrating model-based and model-free RL has also been studied independently of planning \citep{pong2018temporal, janner2019trust}. In contrast, in L2E the model is translated by a planner into long-horizon plans.

In the experiments section, we compare L2E against two representative examples from the literature mentioned above.
The first is using a plan to identify subgoals that are then pursued by an RL agent, as done in \citet{faust2018prm}.
The second is executing the plan using an inverse model, similar to the approach in \citet{christiano2016transfer}.
These two baselines and L2E can be seen as representatives of a continuum: \citet{christiano2016transfer} follow the plan very closely, trying to imitate the planner at each time step. \citet{faust2018prm} relax this requirement and only train the agent to reach intermediate goals.
Finally, in L2E, the agent is free to deviate arbitrarily from the plan (although it is biased to stay close), as long as it reaches the goal.
We find that L2E results in significantly higher success rates when compared against both baselines.

\section{Background}
\label{sec:background}
\subsection{Goal-Conditioned MDPs and RL}
\label{sec:background_mdp}
We consider settings that can be described as discrete-time MDPs $M = \langle \sS, \sA, T, \gamma, R, P_S \rangle$.
$\sS$ and $\sA$ denote the set of all possible states and actions, respectively.
$T: \sS \times \sA \times \sS \rightarrow \mathbb{R}_0^+$ is the transition probability (density); $T(s'|s,a)$ is the probability of the next state being $s'$ if the current state is $s$ and $a$ is chosen as the action.
The agent receives a real-valued reward $R(s,a,s')$ after each transition.
Immediate and future rewards are traded off by the discount factor $\gamma \in [0,1)$.
$P_S: \sS \rightarrow \mathbb{R}_0^+$ is the initial state distribution.

The goal of RL is to learn an optimal policy $\pi^*: \sS \times \sA \rightarrow \mathbb{R}_0^+$ that maximizes the expected discounted return.
In other words, RL algorithms generally try to find
\begin{align}
        \label{eq:rl_problem}
        \pi^* = \argmax_{\pi} \sum_{t=0}^\infty \gamma^t \E_{s_{t+1} \sim T(\cdot\mid s_t,a_t),\, a_{t} \sim \pi(\cdot\mid s_t), s_0 \sim P_S}\left[ R(s_t,a_t,s_{t+1})  \right]
\end{align}
from collected transition and reward data $D = \{(s_i,a_i,r_i,s_i')\}_{i=0}^n$.
More specifically for this work, we are interested in applications in robotics, where both $\sS$ and $\sA$ are typically continuous.
There exists a wide range of algorithms for this case. For the experiments in this paper, soft actor-critic (SAC) \citep{haarnoja2018soft} is used.

In a goal-conditioned MDP $M_G = \langle \sS, \sG, \sA, T, \gamma, R_G, P_S, P_G \rangle$, the reward function $R_G(s,a,s',g)$ has an additional input parameter, the goal $g \in \sG$.
Here, $P_G: \sG  \rightarrow \mathbb{R}_0^+$ is the distribution of goals.
The optimal goal-conditioned policy $\pi^*_G$ acts optimally with respect to any of these goals.

\subsection{Final-Volume-Preserving Reward Shaping}
\label{sec:fvrs}
We use approximate plans as an additional source of information for the RL agent.
For sparse-reward goal-driven MDPs, FV-RS \citep{schubert2021planbased} offers an efficient way to include additional information by adding an additional term
\begin{align}
  R(s,a,s') \rightarrow R_\text{FV}(s,a,s') = R(s,a,s') + F_\text{FV}(s,a,s')
\end{align}
to the reward function, accelerating exploration.
In general, the optimal policy $\pi^*$ corresponding to the original MDP and the optimal policy $\pi^*_\text{FV}$ corresponding to the shaped MDP will be different.
FV-RS however restricts the allowed modifications $F_\text{FV}(s,a,s')$ in such a way that after finite time, the optimally controlled agent ends up in a subset of the volume in which it would have ended up without shaping.
As a result, external information can be made available for the RL algorithm without changing the long-term behavior of the resulting optimal policy.

Specifically in the present work, we consider goal-conditioned MDPs in which the goal-conditioned reward $R_G$ of the underlying MDP is either $1$, if the goal is reached, or $0$ everywhere else.
We further assume that the L2E agent is given an external plan $p$, represented as an intended trajectory $p=(p_1, p_2, \dots)$ in state space.
We intend to reward the agent for staying close to the plan, and for advancing towards the goal along the plan.
A natural way of achieving this is to use a plan-based shaping reward \citep{schubert2021planbased}.
The single-plan shaping function introduced there can be generalized to the multi-plan setting in the present work in the following way:
\begin{align}
  \label{eq:guassian_fvrs_shaping_function}
  F_\text{FV}(s,a,s',p) = \frac{1-R_G(s,a,s',f(p))}{2} \frac{k(s)+1}{L} \exp\left(-\frac{d^2(s,p_{k(s)})}{2 \sigma^2} \right)
\end{align}
Here, $f(p)$ denotes the goal that $p$ leads to, $\sigma\in (0, \infty)$, $k(s) = \argmin_i(d(p_i,s))$, and $d(\cdot,\cdot)$ is a measure of distance in state space.
For the pushing experiments discussed in this work, $d(\cdot,\cdot)$ is the euclidean distance in state space ignoring the coordinates corresponding to the orientation of the box.
The first term in eq. (3) ensures that the assigned shaping reward $F_\text{FV}$ is always smaller than the maximum environment reward (at most $1/2$), and that if the binary environment reward is 1, no shaping reward is assigned. The second term rewards the agent for advancing towards the goal along the plan, and the third term rewards the agent for staying close to the plan.
For a sufficiently high discount factor $\gamma$, $F_\text{FV}$ is final-volume preserving, meaning that the long-term behavior of the optimal agent is unchanged.

\section{Learning to Execute}
\label{sec:method_learning_to_execute}
L2E considers goal-conditioned MDPs $M_G$ (see \secref{sec:background_mdp}), for which an approximate planner $\Omega$ is available.
L2E uses FV-RS to construct a corresponding plan-conditioned MDP $M_P$ from a goal-conditioned MDP $M_G$ and a planner $\Omega$.
In the following sections \ref{sec:plan_mdp} to \ref{sec:replay_strategy}, we introduce our notion of a plan-conditioned MDP $M_P$ and describe the components of the L2E algorithm.
We then summarize the L2E algorithm in \secref{sec:l2ealg}.

\subsection{Plan-Conditioned MDPs}
\label{sec:plan_mdp}
Plans are provided by a model-based planner, which can be described as a distribution $\Omega: \sP \times \sS \times \sG  \rightarrow \mathbb{R}_0^+$ over a set of plans $\mathbb{P}$.
Given an initial state and a goal, $\Omega(p|s,g)$ is the probability that the planner outputs $p$ as a possible plan of how to achieve $g$ from state $s$.
The distinction between goals and plans is that plans are conditional on \emph{both a goal and an initial state}.
Therefore, both initial state and goal can be inferred using the plan only.

In a plan-conditioned MDP $M_P = \langle \sS, \sP, \sA, T, \gamma, R_P, P_S, P_P \rangle$, a plan $p \in \sP$ is given to the reward function $R_P(s,a,s',p)$ as an additional input parameter. $P_P:\sP  \rightarrow \mathbb{R}_0^+$ is the distribution of plans.
The optimal plan-conditioned policy $\pi_P^*$ behaves optimally with respect to any of these plans, creating a distribution $\pi_P^*(\cdot \mid s, p)$ over actions that is conditional on the current state and the current plan.

\subsection{Constructing the Plan-Conditioned MDP}
\label{sec:construct_mdp}
We use FV-RS to shape the reward function $R_G$ of the original goal-conditioned MDP $M_G = \langle \sS, \sG, \sA, T, \gamma, R_G, P_S, P_G \rangle$ with a plan-dependent term $F_\text{FV}(s,a,s',p)$ (see \eqref{eq:guassian_fvrs_shaping_function})
\begin{align}
  R_G(s,a,s',g) \rightarrow R^\text{FV}_G(s,a,s',g,p) = R_G(s,a,s',g) + F_\text{FV}(s,a,s',p) \quad \text{.}
\end{align}
We call $g = f(p)$ the goal for which the plan p was created. If a planner $\Omega$ should be such that g can not be recovered from the resulting plan $p\sim\Omega(.|s,g)$, we can always construct a new $\tilde{p}\sim\tilde{\Omega}$ such that $\tilde{p} = [p, g]$.
Since now $g$ can be recovered from $\tilde{p}$ deterministically, we can assume that $f$ always exists without loss of generality.
We can interpret the shaped reward function
\begin{align}
  R_P(s,a,s',p) = R^\text{FV}_G(s,a,s',f(p),p)
\end{align}
as a plan-conditioned reward function of a plan-conditioned MDP $M_P = \langle \sS, \sG, \sA, T, \gamma, R_P, P_P \rangle$.
The distribution over initial states and plans $P_P$ of $M_P$ is still missing, and can be constructed as
\begin{align}
  P_P(s,p) = \int \Omega(p|s,g) P_S(s) P_G(g) \mathrm{d}g \quad \text{.}
\end{align}
In practice, $P_P$ can be sampled from by first sampling $s\sim P_S\text{, } g\sim P_G$ and then subsequently sampling $p\sim\Omega(\cdot|s,g)$.

Thus, we have constructed a plan-conditioned MDP $M_P$ by combining a goal-conditioned MDP $M_G$ with an approximate planner $\Omega$ and a FV-RS shaping function $F_\text{FV}$.
For reference later in this paper, we write as a shorthand notation $M_P=\mathcal{C}(M_G, \Omega, F_\text{FV})$.
Furthermore, we will refer to $M_P$ as the corresponding plan-conditioned MDP to $M_G$ and vice versa.

In contrast to potential-based reward shaping \citep{ng1999policy}, FV-RS does not leave the optimal policy invariant.
As a result, generally $\exists p \in \sP: \pi_G^*(\cdot|\cdot ,f(p)) \not\equiv \pi_P^*(\cdot|\cdot,p)$.
In words, the optimal policy of $M_P$ and the optimal policy of $M_G$ will not result in identical behavior.
In fact, while $\pi_G^*(\cdot|\cdot , g)$ learns one policy for each goal $g$, $\pi_P^*(\cdot|\cdot , p)$ can learn different behavior for each plan in the set of plans $\{p \in \sP \mid f(p)=g\}$ leading towards the same goal $g$.

\subsection{Plan Replay Strategy}
\label{sec:replay_strategy}
In order to efficiently learn a universal plan-conditioned L2E policy, the reward for experienced episodes is evaluated with respect to many different plans.
In HER \citep{andrychowicz2017}, it is assumed that each state $s \in \sS$ can be assigned an achieved goal.
Recorded episodes are then replayed with respect to goals that were achieved during the episode, i.e.~the recorded transitions are re-evaluated with respect to these goals.
This ensures that the recorded transitions were successful in reaching the replayed goals, resulting in highly informative data.

In L2E, transitions are replayed with respect to plans.
However, there is no meaningful relation between each state $s \in \sS$ and a unique ``achieved plan''.
Therefore, the L2E agent replays transitions with past plans that were recorded at some point during training and were stored in its replay buffer $D$.
The replay plans are chosen according to a plan replay strategy $S_n$.

A plan replay strategy $S_n$ provides a distribution over $n$ replay plans, conditioned on the replay buffer $D$ and the buffer containing the current episode $D_\text{ep}$ (see \algref{alg:l2e} for a definition of $D$ and $D_\text{ep}$).
For replay, $n$ plans are sampled according to this strategy $\{p_1,\dots,p_n\} \sim S_n(\cdot \mid D_\text{ep}, D)$.
We consider two types of replay strategies.
Uniform replay $S^\text{uni}_n$ samples $n$ unique plans uniformly from the replay buffer $D$.
Reward-biased replay $S^\text{bias}_{nm}$ first uniformly samples $m$ unique plans from the replay buffer $D$, and then returns the $n$ plans $p_i$ that would have resulted in the highest sum of rewards $\sum_{(s_k, a_k, s_k') \in D_\text{ep}} R_P(s_k,a_k,s_k',p_i)$ for the episode stored in $D_\text{ep}$.
The idea behind using reward-biased replay is to bias the replay towards transitions resulting in higher reward.

\begin{algorithm}
  \label{alg:l2e}
  \SetKwInOut{Input}{Input}
  \SetKwInOut{Output}{Output}
  \Input{Goal-conditioned MDP $M_G$, approximate planner $\Omega$, FV-RS shaping function $F_\text{FV}$, plan replay strategy $S_n$, off-policy RL Algorithm $\mathcal{A}$}
  \Output{Universal plan-conditioned optimal policy $\pi^*_P$ for the corresponding plan-conditioned MDP $M_P=\mathcal{C}(M_G, \Omega, F_\text{FV})$}
  Construct plan-conditioned MDP $M_P=\mathcal{C}(M_G, \Omega, F_\text{FV})$ as detailed in \secref{sec:construct_mdp}\;
  Initialize replay buffer $D \leftarrow \{ \}$\;
  \While{$\pi^*_P$ not converged}{
   Initialize episode buffer $D_\text{ep} \leftarrow \{ \}$\;
   Sample initial state and goal $(s_0, g)\sim P_G$\;
   Sample plan $p\sim\Omega(\cdot|s_0,g)$\;
   $s \leftarrow s_0$\;
   \While{Episode not done}{
    Sample action $a\sim \pi^*_P(\cdot\mid s,p)$\;
    Sample transition $s' \sim T(\cdot\mid s,a)$\;
    Collect shaped reward $r \leftarrow R_P(s,a,s',p)$\;
    Add to episode buffer $D_\text{ep} \leftarrow D_\text{ep} \cup \{(s,a,r,s',p)\}$\;
    $s\leftarrow s'$\;
   }
  Add episode to replay buffer $D \leftarrow D \cup D_\text{ep} $\;
  Get replay plans $\{p_1, \dots,p_n \} \sim S_n(\cdot \mid D_\text{ep}, D)$\;
  \For(){
    $p_\text{replay}$ in $p_1, \dots, p_n$ 
  }{
    \For(){
      $(s,a,r,s',p)$ in $D_\text{ep}$
    }{
      Calculate replay reward $r_\text{replay} \leftarrow R_P(s,a,s',p_\text{replay})$\;
      Add replayed transition to buffer $D \leftarrow D \cup \{(s,a,r_\text{replay},s',p_\text{replay})\}$\;
    }
  }
  Update policy using off-policy RL algorithm $\pi^*_P \leftarrow \mathcal{A}(\pi^*_P, D)$
  }
  \caption{Learning to Execute (L2E)}
\end{algorithm}

\subsection{L2E Algorithm}
\label{sec:l2ealg}
The L2E algorithm is outlined in \algref{alg:l2e}.
First, the corresponding plan-conditioned MDP $M_P=\mathcal{C}(M_G, \Omega, F_\text{FV})$ is constructed from the original goal-conditioned MDP $M_G$, the planner $\Omega$ and the shaping function $F_\text{FV}$ as described in \secref{sec:construct_mdp}.
The agent acts in the environment trying to follow one randomly sampled plan per episode.
The episode is then added to the replay buffer, along with data from episode replays with respect to other plans.
These other plans are sampled from the replay buffer according to the replay strategy $S_n$.
A generic off-policy RL algorithm is used to update the agent using the replay buffer.
This process is repeated until convergence.

We would like to emphasize that the L2E algorithm is agnostic to the exact type of off-policy RL algorithm.
By combining state and plan into a ``super state'' for the purpose of passing the replay buffer to the off-policy RL algorithm, L2E can be interfaced with any off-the-shelf implementation.

\section{Experiments}
\label{sec:experiments}
We evaluate the L2E agent against several baselines using two simulated robotic manipulation tasks, namely a pushing task and an obstacle avoidance task.
These two environments are chosen to compare different approaches on a variety of challenges.
While the pushing task can be seen as an open-source version of the opanAI gym FetchPush-v1 task \citep{brockman2016openai}, the obstacle task was chosen to represent robotic manipulation tasks with segmented state spaces.
This allows us to discuss limitations of exploration in such environments as well.

A video of the experiments is available in the supplementary material.
The complete code to fully reproduce the figures in this paper from scratch can be found at \href{https://github.com/ischubert/l2e}{github.com/ischubert/l2e} and in the supplementary material.
This includes the implementation of the environments, the implementation of the L2E agents and the baselines, and the specific code used for the experiments in this paper.

The experiments section is structured as follows.
In \secref{sec:env_and_planner} we discuss the environments and planners that are used in the experiments.
We briefly introduce the plan embedding used for the L2E agent in \secref{sec:plan_encoding}, additional experiments on this can be found in \secref{sec:experiments_plan_encoding}
In \secref{sec:baselines} we introduce the baselines against which we compare our method.
In \secref{sec:results} we discuss our experimental results.
Implementation details of the L2E agent are given in \secref{sec:method}

\subsection{Environments and Planners}
\label{sec:env_and_planner}
\begin{figure}
  \centering
  \subfloat[\label{fig:pushing_setup}]{%
          \stackinset{r}{-0.0pt}{t}{-5pt}{
            \includegraphics[width=0.3\linewidth]{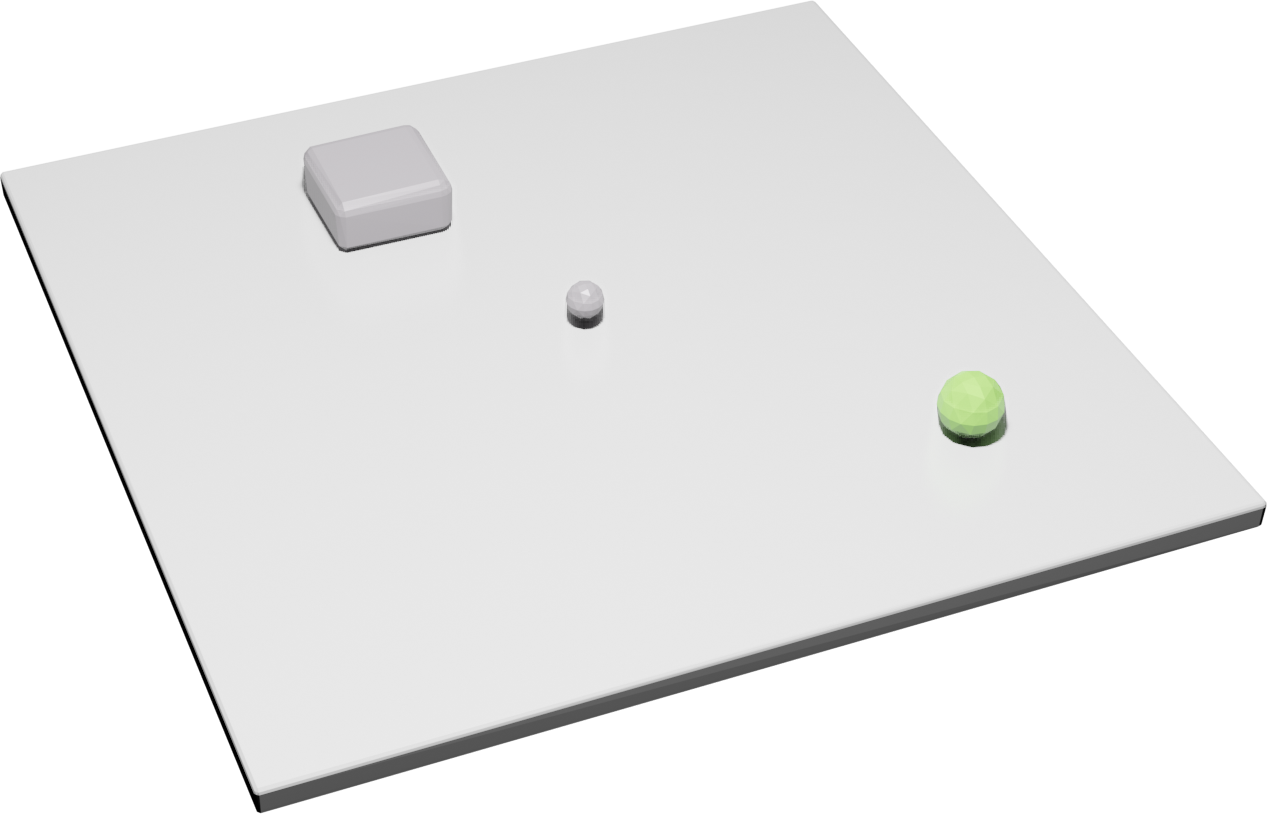}
          }{
            \includegraphics[width=0.514\linewidth]{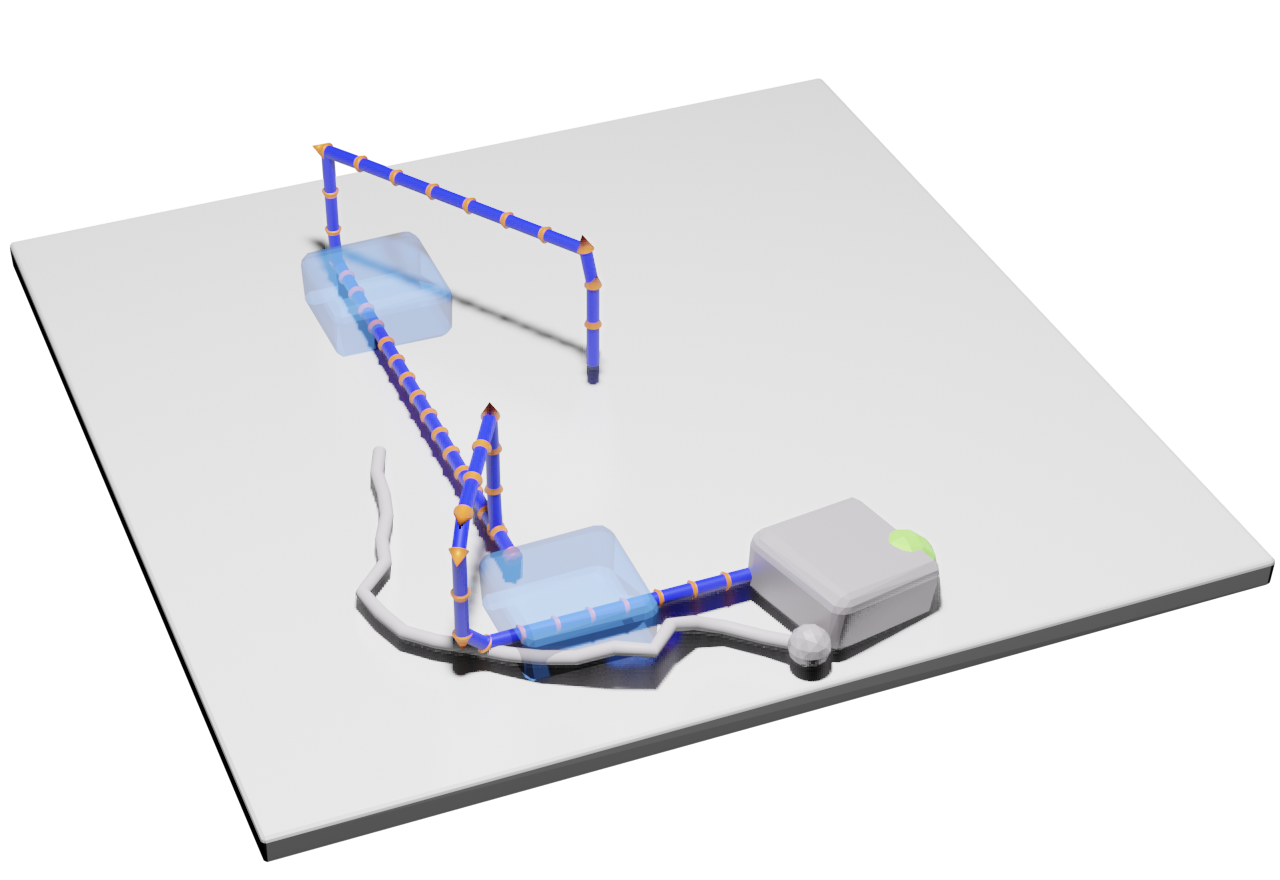}
          }
  }
  \subfloat[\label{fig:results_pushing}]{%
          \includegraphics[width=0.485\linewidth]{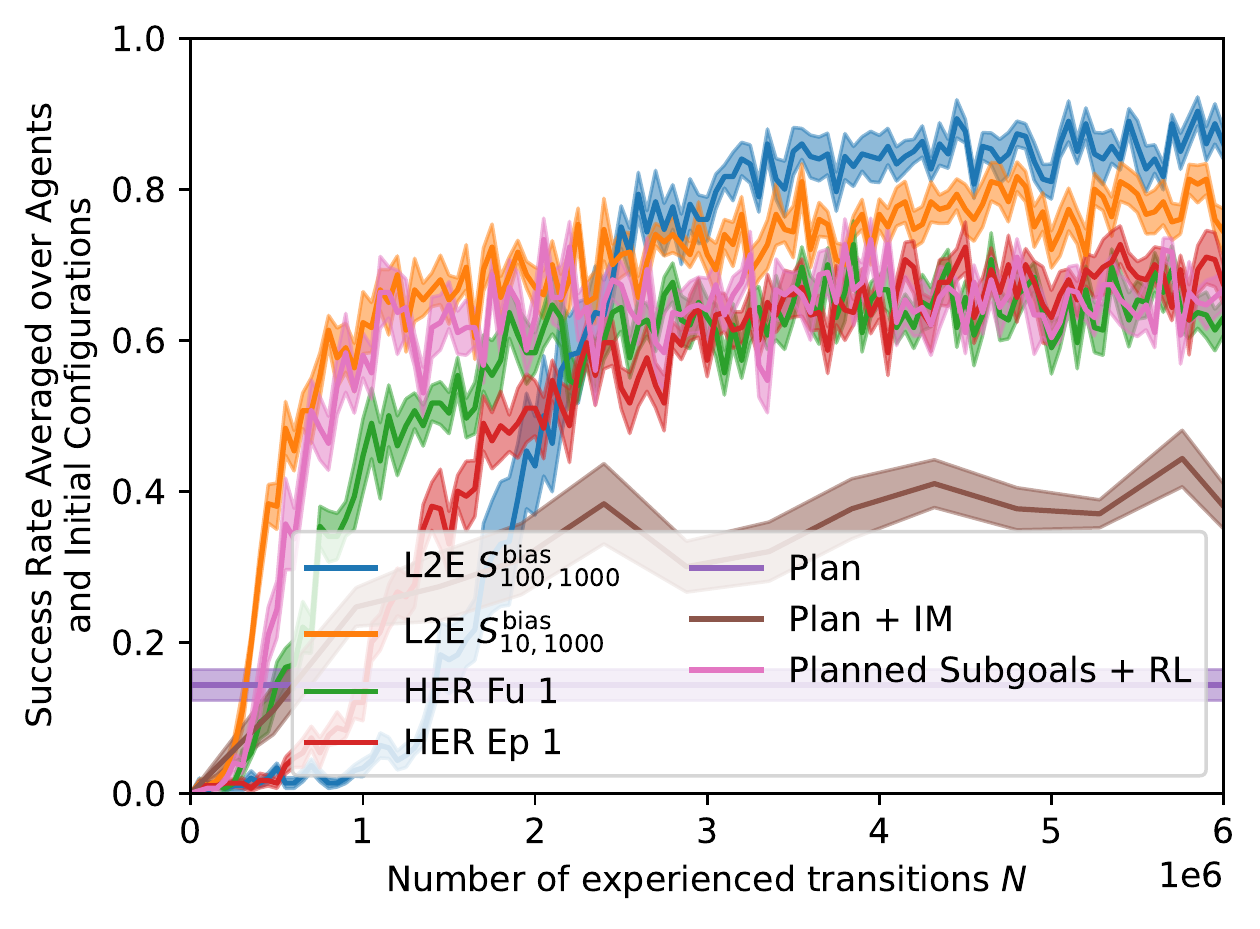}
  }\\
  \subfloat[\label{fig:obstacle_setup}]{%
        \stackinset{r}{-0.0pt}{t}{-10pt}{
          \includegraphics[width=0.3\linewidth]{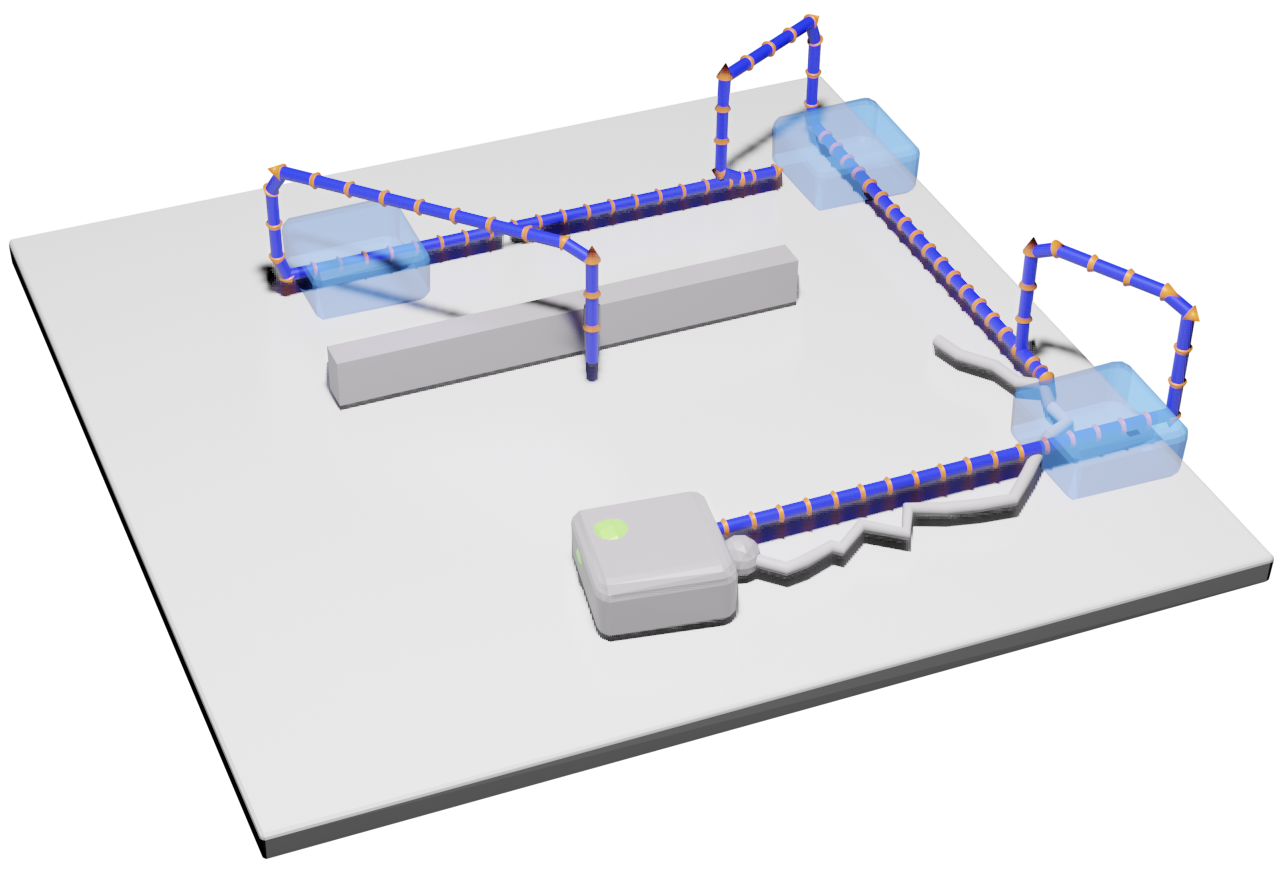}
        }{
          \includegraphics[width=0.514\linewidth]{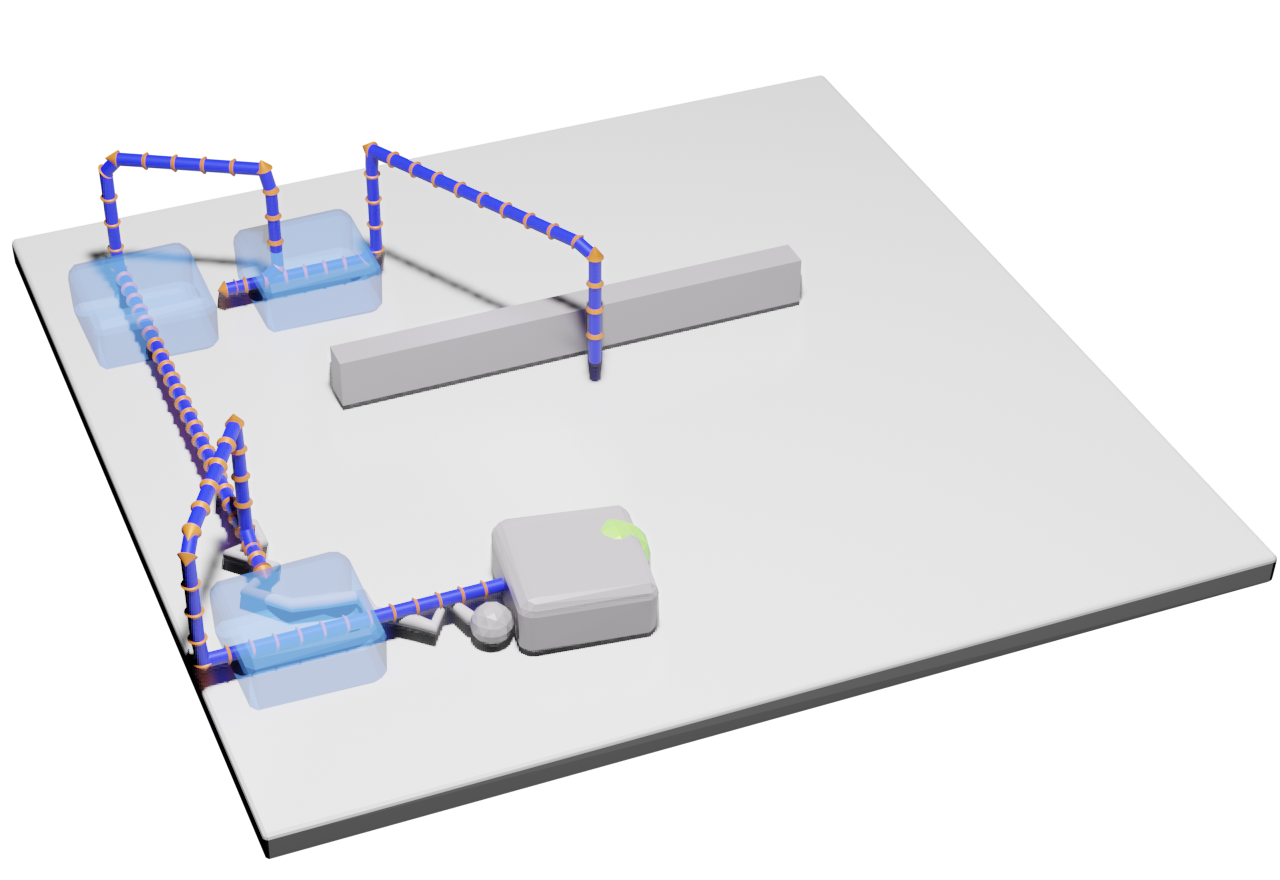}
        }
  }
  \subfloat[\label{fig:results_obstacle}]{%
          \includegraphics[width=0.485\linewidth]{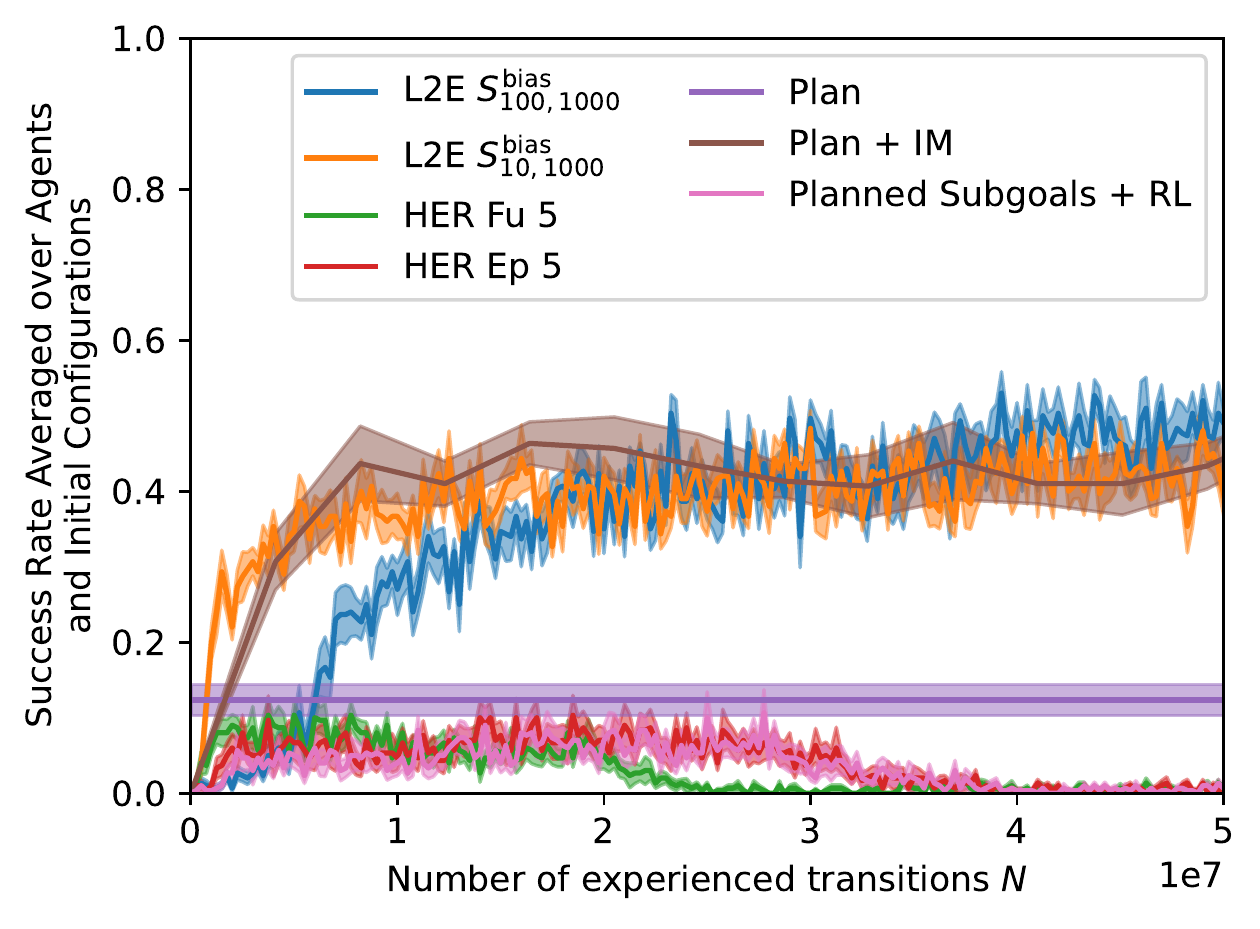}
  }
  \caption{
          Results for the basic pushing environment (figures (a) and (b)) and the obstacle pushing environment (figures(c) and (d)).
          Lines represent averages over $N$ independently trained agents times $M$ test rollouts per agent and time step, the colored areas indicate the confidence intervals (standard deviations of the mean).
          Renderings created with open-source code by \citet{BpyScripting}.
          (a) The simulated pushing setup (inset). The spherical end effector tries to push the box to the green target. Once the box is pushed off the table, it can not be recovered.
          An approximate planner creates simple manhattan-like plans (blue line and transparent boxes in large figure; the end effector is planned to change height to push the box from the other side).
          The path of the box is planned fully, but only shown at some time steps.
          Recent end effector movement is indicated in grey.
          Learned behavior can differ significantly from the original plan.
          (b) Comparison to the baselines introduced in \secref{sec:baselines}.
          L2E shows significantly higher sample efficiency and success rates.
          See \secref{sec:longer_run_l2e_100_1000_basic_pushing} for a longer run of L2E ${S^\mathrm{bias}}_{100, 1000}$.
          (c) In the simulated obstacle setup, an additional obstacle is added. In this environment, the planner can create (and the RL agent learns to execute) multiple plans from the same initial configuration to the same goal (inset shows alternative plan).
          (d) Comparison to the baselines introduced in \secref{sec:baselines}.
          L2E performs significantly better than the pure learning HER baselines, the pure planning baseline ("Plan"), and the “Planned Subgoals + RL” baseline.
          While using an inverse model is initially more efficient, L2E achieves significantly better results if given enough data.
  }
  \label{fig:results}
\end{figure}

\Figref{fig:pushing_setup} and \Figref{fig:obstacle_setup} show renderings of the basic pushing environment and obstacle pushing environment, respectively.
We use the open-source Nvidia PhysX engine \citep{physxengine} to simulate a box of size $0.4 \times 0.4$ being pushed on a table of size $3 \times 3$ by a spherical end effector of radius $0.06$.
The 10D state space of both the goal-conditioned MDP $M_G$ and the corresponding plan-conditioned MDP $M_P$ consists of the 3D position of the end effector, the 3D position of the box, and the 4D quaternion for the orientation of the box.
The agent controls the 3D velocity of the end effector.
The maximum velocity in any direction is $0.1$ per time step.
The end effector movement resulting from the agent's actions is slightly distorted by random noise.
In the obstacle pushing environment, the agent additionally has to evade an obstacle in the middle of the table.

In the goal-conditioned MDP $M_G$, each goal is represented as a desired 2D box position on the table.
The goal-dependent sparse reward function $R_G$ is $1$ if the box is within $0.1$ of this desired goal, and $0$ if not. The initial state-goal distribution $P_G$ is uniform across the table for the non-colliding box position and goal position.
The end effector is always initialized at the origin and the box is always initialized with a fixed orientation parallel to the table.

For the basic pushing environment, we use a crude manhattan-like planner $\Omega$ that deterministically outputs plans consisting of two separate contacts leading the initial state to the goal as shown in \Figref{fig:pushing_setup}. For the obstacle pushing environment, plans consist of four contacts, corresponding to an additional intermediate box position which is chosen at random (see \Figref{fig:obstacle_setup}).
Thus, the agent learns to execute an infinite number of plans for each combination of start and goal.

Plans are represented as a trajectory of length $50$ for the basic pushing environment and $100$ for the obstacle pushing environment, consisting of 6D elements representing end effector position and box position.
For the basic pushing environment, we additionally report results for less dense plans in \secref{sec:experiments_plan_density}.
The orientation of the box is not specified in the plans.
We construct the plan-conditioned MDP $M_P$ as described in \secref{sec:construct_mdp}, using this planner and the FV-RS function in \eqref{eq:guassian_fvrs_shaping_function}. We use the width parameter $\sigma=0.5$ throughout the experiments.

\subsection{Plan Encoding}
\label{sec:plan_encoding}
The plans $p$ are embedded before they are provided to the policy. A plan encoding is an injective function $\phi: \mathbb{P} \rightarrow \mathbb{C}$ from the set of plans $\mathbb{P}$ to a latent space $\mathbb{C}$.
If $\mathbb{P}$ is a manifold in some high-dimensional space, the dimensionality of the latent space must be at least as high as the dimensionality of the manifold.
Since $\mathbb{P}$ is task-dependent, the encoding will be task-dependent as well. For the basic pushing environment (\Figref{fig:pushing_setup}), $\mathbb{P}$ is a 4D manifold (since the plans only depend on the initial and final 2D box positions).
For the obstacle task (\Figref{fig:obstacle_setup}), $\mathbb{P}$ is a 6D-manifold (since the plans depend on one intermediate box position as well).

In the experiments discussed in the present work, we encode plans analytically using box positions as described above.
We experimentally compare this with either learning the encoding or not using any encoding at all in section \ref{sec:experiments_plan_encoding}.

\subsection{Baselines}
\label{sec:baselines}
We compare L2E against (1) direct plan execution, (2) plan execution with an inverse dynamics model, (3) using RL to reach subgoals, and (4) HER.
We describe these baselines in detail in \secref{sec:baselines_detail}.

\subsection{Results}
\label{sec:results}
Both at training and evaluation time, we run episodes of length $250$.
For each method $q$ (i.e., L2E and all baselines), we independently train $A=10$ agents.
After $N$ environment transitions, we evaluate the agents.
We reset to random initial positions and goals/plans and run the experiment until the goal is reached or until the episode ends.
We repeat this process $M=30$ times for each agent, and store whether the rollout was successful in reaching the goal.
We denote the result of the $m$-th evaluation of the $a$-th agent for method $q$, evaluated after learning for $N$ environment transitions, as $\mathcal{F}^{(q)}_{am}(N)$.

As can be seen from the video given in the supplementary material, even though the L2E agent uses plan information as a guiding bias during exploration, and is encouraged to stay close to the plan by the shaping reward, it can also learn to deviate considerably from the plan if closely following it will be suboptimal for reaching the goal fast.
For example, while the simple planner (see \Figref{fig:pushing_setup} and \Figref{fig:obstacle_setup}) suggests to re-establish the contact during the sequence, the L2E agent almost always moves and turns the box using a single contact.

\subsubsection{Basic Pushing Environment}
\label{sec:basic_pushing_environment_results}
To allow for a fair comparison, we spent a considerable amount of effort to optimize the HER replay strategy as well as the L2E strategy.
Details on this are given in \secref{sec:her_and_l2e_strategy}.

The results for the pushing setup are summarized in \Figref{fig:results_pushing}.
We observe that both L2E versions outperform all baselines in terms of the asymptotical performance.
L2E with biased replay strategy $S_{10, 1000}$ exhibits a high sample efficiency especially in the beginning, resulting in success rates significantly higher than $50\%$ after $4000$ episode rollouts or $1$ Million time steps.
Directly executing the plan results in very low success rates of significantly less than $20\%$ on average.
Executing the plan with an inverse model (IM) still shows significantly worse long-term performance than the RL methods.
HER results in better policies than the IM baselines, but is relatively data hungry.
This can be improved slightly if the HER agent is only used to reach subgoals given by the planner.

Pushing is a challenging interaction that requires reasoning for several time steps ahead.
A typical failure mode of the IM baseline (see also videos) is that the box moves away from the intended trajectory too much, so that the agent is not able to correct for it within one time step.
In contrast, the L2E agent learns to deviate from the planned trajectory if this is required to reach the goal.

We find that L2E, combining a model-based planner and a universal plan-conditioned policy, outperforms our baselines that are pure planning or pure learning approaches.
In addition, L2E outperforms the two baselines that also combine learning and planning.

\subsubsection{Obstacle Pushing Environment}
L2E performs significantly better than the pure learning HER baselines, the pure planning baseline ("Plan"), and the “Planned Subgoals + RL” baseline.
While using an inverse model is initially more efficient, L2E achieves significantly better results if given enough data.

Comparing the basic pushing environment (\secref{sec:basic_pushing_environment_results}) to the obstacle environment, L2E learns slower in the latter.
This is in part due to the higher dimensionality of the latent space of plan encodings (see also \secref{sec:plan_encoding}), posing a more challenging learning problem to the L2E agent.
In contrast, the "Plan+IM" baseline is independent of the size of the plan space, and performs comparably to the experimental setting in the original version.

The obstacle in the middle segments the state space into two parts.
In order to move from one side to the other, an agent already has to be able to reason about long-term results of its actions.
As evidenced by the results for HER, this poses a significant challenge for pure RL.
Incorporating planner knowledge helps the agent to overcome this chicken-and-egg problem.

\section{Discussion}
\label{sec:discussion}
Learning plan-dependent policies as opposed to goal-dependent policies has the additional advantage that the former can learn to execute multiple plans that lead from the same initial state to the same goal, as shown in the obstacle environment.
Thus, the policy learns multiple strategies to achieve the same outcome.
In principle, this allows it to adapt to changed scenarios where some of these strategies become infeasible.
If, e.g., the environment changes, it suffices to only update the planner’s crude model of the environment so that it creates plans that are feasible again.
These can then be directly fed into the policy without retraining.
We explore this possibility in \secref{sec:moving_obstacles},
using a simple 2D maze environment with moving obstacles.
We find that the plan-conditioned L2E policy consistently achieves 90\% success rate in this quickly changing environment, while the goal-conditioned HER policy does not improve beyond 60\% success rate.

We used rather simple plans to support the RL agent during training,
and demonstrated that these are already sufficient to significantly speed up learning in our experiments.
In fact we demonstrate in \secref{sec:experiments_plan_density} that in the basic pushing example, the L2E agent is very robust against plans of even lower quality.
Using simple plans enabled us to use an analytical encoding;   for very complex scenarios it might be beneficial to learn the encoding using an auxiliary objective (see, e.g., \citet{co2018self}).
We present results on using a variational autoencoder (VAE) in \secref{sec:experiments_plan_encoding}.

The use of FV-RS biases the RL agent towards following the plan. While it was shown in the experiments that the RL agent can learn to deviate from the plan, plans that are globally misleading can act as a distraction to the agent. In the present work, it is assumed that plans can be used to guide the agent during learning, increasing sample efficiency. Independently of the specific method used to achieve this, misleading plans will always break this assumption.

Comparing the basic pushing environment to the obstacle pushing environment, the amount of data needed for learning a plan-conditioned policy clearly depends on the size of the plan spaces that are considered.
For very large plan spaces, more data will be needed to master the task.
Still, including planner information into the learning process makes a decisive difference, as demonstrated by the relative performance of L2E and HER in the obstacle example.

While SAC was used for the experiments in \secref{sec:experiments}, L2E can be used in combination with any off-policy RL algorithm. L2E reformulates a goal-conditioned MDP as a plan-conditioned MDP, and provides a replay strategy to efficiently solve the latter.
It is agnostic to how this data is then used by the RL agent.

The specific FV-RS shaping function used in this work applies to MDPs with sparse rewards.
We focused on this since sparse rewards are common in robotic manipulation.
In addition, they often present notoriously hard exploration tasks, making external plan-based information as used by L2E particularly useful.
However, FV-RS in general is not restricted to sparse-reward settings, and by using a different shaping function, L2E could be applied in other settings as well.

Apart from FV-RS, there are alternative schemes of reward shaping such as potential-based reward shaping (PB-RS) \cite{ng1999policy}.
In principle, these could also be used to increase the sample efficiency of the RL agent.
We chose FV-RS for two reasons.
First, in the original paper \cite{schubert2021planbased}, it was demonstrated that FV-RS leads to significantly higher sample efficiency than PB-RS.
Second, since PB-RS leaves the optimal policy invariant, the behavior of the fully converged policy trained with PB-RS will only be goal-dependent, and not depend on the rest of the plan.

The original HER paper \citep{andrychowicz2017} considers the use of a simple form of reward shaping in combination with HER as well.
It is found that reward shaping dramatically reduces the performance of HER in a robotic pushing task.
In the present work, we show in contrast that including plan information using FV-RS shaping improves the performance of RL in a similar task.
A possible explanation to reconciliate these seemingly contradictory results is already offered by \citet{andrychowicz2017}:
While simple domain-agnostic shaping functions can be a distraction for the RL agent, domain-specific reward shaping functions can be beneficial.
This view is supported, e.g., by similar results by \citet{popov2017data}.
\citet{andrychowicz2017} state that however ``designing such shaped rewards requires a lot of domain knowledge''.
In this context, one could view L2E as an automated way to extract such domain-specific knowledge from model-based planners and make it available.
We specifically believe that L2E can be useful in robotic manipulation tasks, where domain knowledge is in fact readily available in many cases.
Here, L2E offers a way to exploit this.

\section{Conclusion}
\label{sec:conclusion}
We introduced L2E, an algorithm that links RL and model-based planning using FV-RS.
RL generally results in well-performing policies but needs large amounts of data, while model-based planning is data-efficient but does not always result in successful policies. 
By combining the two, L2E seeks to exploit the strengths of both approaches.
We demonstrated that L2E in fact shows both higher sample efficiency when compared to purely model-free RL,
and higher success rates when compared to executing plans of a model-based planner.
In addition, L2E also outperformed baseline approaches that combine learning and planning in our experiments.

\begin{ack}
  The authors would like to thank Valentin N Hartmann for stimulating discussions.
  The research has been supported by the International
  Max-Planck Research School for Intelligent Systems (IMPRS-IS), and by the
  German Research Foundation (DFG) under Germany’s Excellence Strategy EXC 2120/1–390831618 ``IntCDC'' and EXC 2002/1–390523135 ``Science of Intelligence''.
\end{ack}
\bibliography{bibfile}
\bibliographystyle{iclr2021_conference}

\newpage
\appendix

\section{Appendix}

\subsection{L2E Implementation}
\label{sec:method}
We use the open-source Stable Baselines3 \citet{stable-baselines3} SAC agent for all experiments.
To allow for a fair comparison, we use the same hyperparameters for the SAC algorithm in all L2E and HER runs.
Both L2E and the HER baseline are trained using rollout episodes of length $250$.

For the actor and critic networks, we use 6 hidden layers with sizes decreasing geometrically from $1024$ to $64$.
We train the networks using $0.0003$ as the learning rate in batches of size $256$.
The replay buffer stores $1\times 10^6$ transitions.
The L2E and HER agents need ca.\,1GB of GPU memory and are trained using an Nvidia RTX-3090 GPU and AMD EPYC 7502 CPU.
For full details on the implementation, please also refer to the code at \href{https://github.com/ischubert/l2e}{github.com/ischubert/l2e} and in the supplementary material.

\subsection{Details on Baselines}
\label{sec:baselines_detail}
We compare L2E against several baselines that we describe in the following.
\subsubsection{Plan Execution}
\label{sec:basline_direct_plan_execution}
As basic baseline, we compare against direct execution of the plan using a simple controller.
Here, at each step the point $p_k$ in the planned trajectory that is closest to the current position $s$ is selected.
To measure the distance, we use $d(\cdot,\cdot)$ introduced in \secref{sec:fvrs}.
The action is then calculated as the delta between the end effector position at this point and the following point $p_{k+1}$ of the trajectory.
We stop the run if either the goal or the time limit of $250$ steps is reached.

\subsubsection{Plan Execution with Inverse Dynamics Model}
We use an inverse dynamics model to transfer the plan-based controller described in \secref{sec:basline_direct_plan_execution} to the simulated environment.
This can be seen as a transfer method as introduced by \citet{christiano2016transfer}.

We learn an inverse dynamics model (IM) $\phi: \sS \times \sS \rightarrow \sA$ from simulation data.
From input pairs of state $s$ and desired next state $s'$, this model learns to predict feasible actions $a=\phi(s,s')$.
To collect data, we reset the environment to random initial states, and then perform random actions.
In order to bias the data collection towards more informative samples, an action leading the end effector directly towards the box is randomly selected at $10\%$ of all time steps.
After $250$ time steps, we reset and repeat the process.
We then train a neural network to predict actions from pairs of state and next state using mean squared error loss.
We use the open-source Pytorch \citep{NEURIPS2019_9015} package.

Evaluation is done in the same way as described in \secref{sec:basline_direct_plan_execution}, with the only difference that at each step we calculate the action $a=\phi(s,p_{k+1})$ using the inverse dynamics model. Since the planned trajectory does not specify the box's orientation, we always assume the desired orientation to be parallel to the table.

\subsubsection{Planned Subgoals and RL}
Here we use the planner to create a sequence of subgoals that can be navigated by an RL agent.
This approach is inspired by PRM-RL \citep{faust2018prm}.

The subgoal is selected as a box position on the plan that is $0.3$ ($1/10$ of the table length) away from the current box position. Once the subgoal is reached (tolerance $0.1$), the next subgoal is selected.

Thus, long-term planning is provided by the planning module.
The RL agent on the other hand is tasked with learning the dynamics of the system to reach relatively short-term goals.
This RL agent is conditioned on subgoals, and trained using HER.
We use the sampling strategies "Future 1" in the basic pushing environment (see \Figref{fig:pushing_setup}), and "Episode 5" in the obstacle pushing environment (see \Figref{fig:obstacle_setup}).
These performed best on the respective environment when used without any planning.

\subsubsection{HER}
We use the open-source Stable Baselines3 \citet{stable-baselines3} implementation of HER to learn universal goal-conditioned policies for the goal-conditioned MDP $M_G$.
To allow for a fair comparison, we run several experiments comparing different HER replay strategies.
As described in \secref{sec:method}, we use SAC with the same hyperparameters both for HER and for L2E.

\subsection{Plan-conditioned Policies in a Constantly Changing Environment}
\label{sec:moving_obstacles}
Since it learns a plan-conditioned policy in contrast to a goal-conditioned policy, the L2E agent can adapt to changed scenarios where some of the initially successful strategies become infeasible.
If, e.g., the environment changes, it suffices to only update the planner’s crude model of the environment so that it creates plans that are feasible again.
These can then directly be fed into the policy without retraining.
This is an advantage that is specific to learning plan-conditioned policies over goal-conditioned policies.
We demonstrate this using a 2D maze toy example with moving obstacles in the following.

\subsubsection{Experimental Setup}
We consider a 2D world of size $1\times 1$.
In a single step, the agent can move to any point within a radius of $0.1$.
This movement is distorted by random noise with maximum absolute value $0.01$.
If an obstacle is in the way or the agent reaches the border, nothing happens.

\begin{figure}[h]
  \centering
  \subfloat[\label{fig:moving_obstacles_env_0}]{%
  \includegraphics[width=0.5\linewidth]{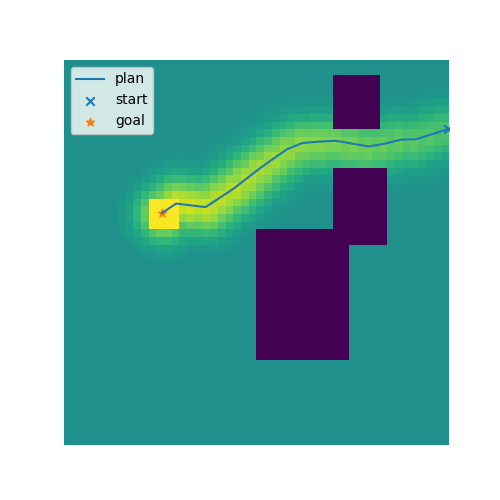}
  }
  \subfloat[\label{fig:moving_obstacles_env_1}]{%
  \includegraphics[width=0.5\linewidth]{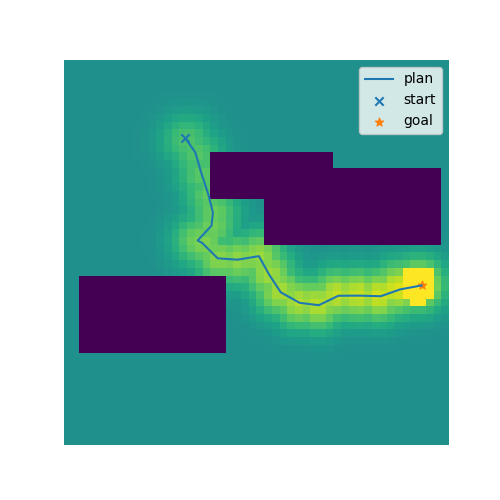}
  }\\
  \subfloat[\label{fig:moving_obstacles_env_2}]{%
  \includegraphics[width=0.5\linewidth]{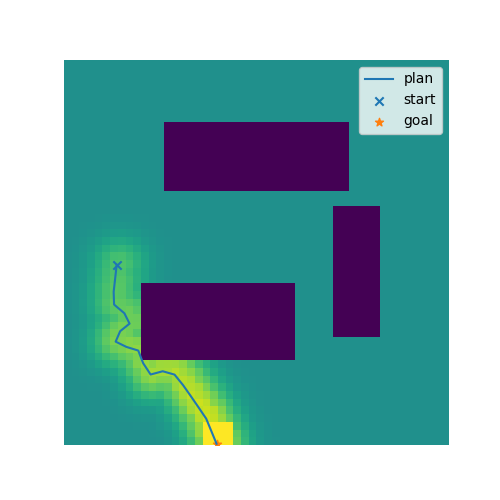}
  }
  \subfloat[\label{fig:moving_obstacles_results}]{%
  \includegraphics[width=0.5\linewidth]{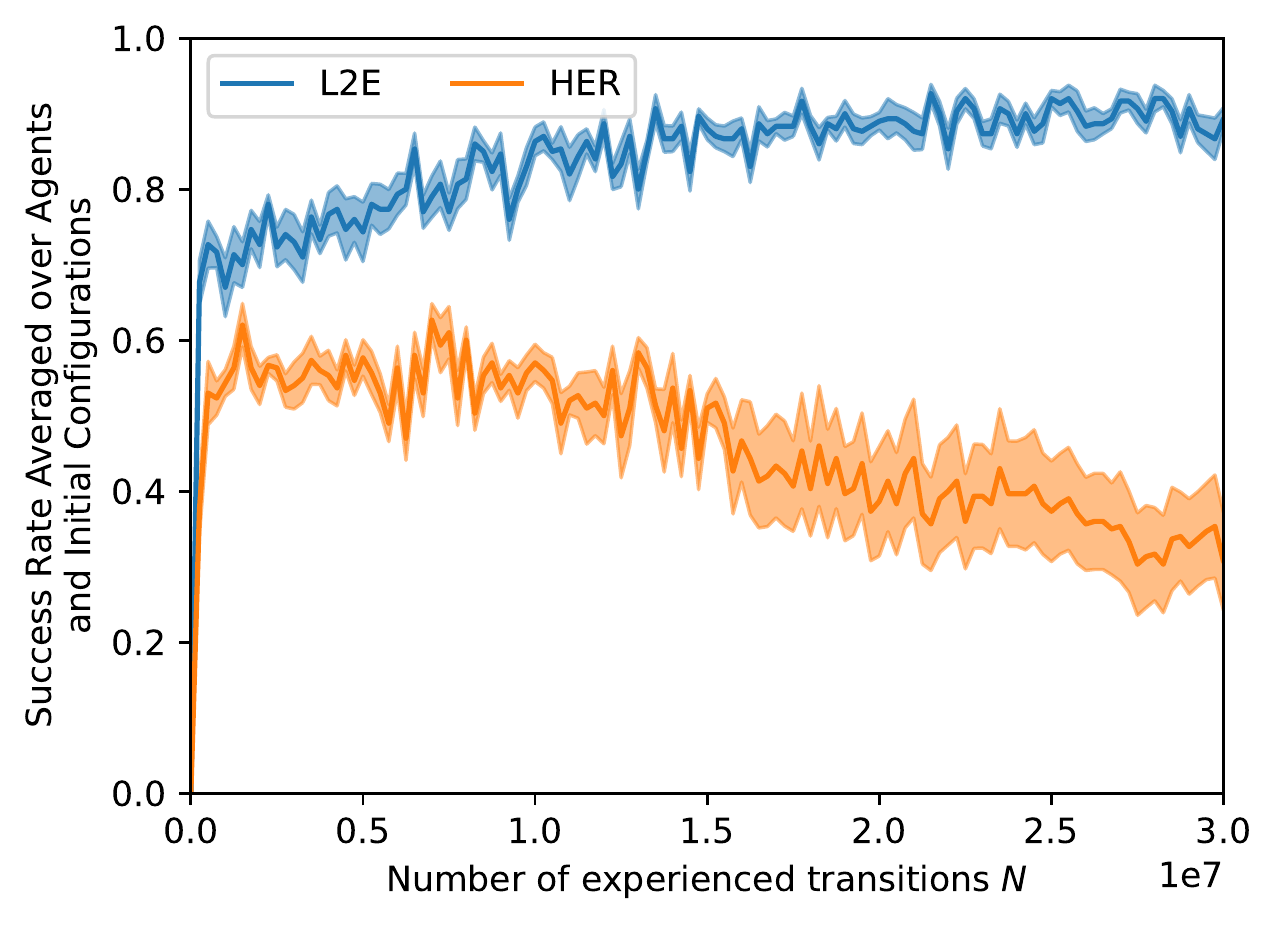}
  }
  \caption{
    (a, b, c) The positions of the obstacles in the 2D maze environment change after each episode.
    Three randomly selected configurations are shown here; obstacles are shown in dark purple.
    For each configuration, a feasible plan from start to goal is created by an RRT planner (we use an implementation by \citet{zeng2019rrt}).
    The yellow heatmap indicates the value of the FV-RS shaping reward created by the L2E agent.
    (d) Since the L2E agent learns to execute plans, it learns to become consistently successful in this environment by following the plan.
    In contrast, the HER agent is unable to effectively generalize to a changed environment.
  }
  \label{fig:moving_obstacles_env}
\end{figure}

At the end of each episode of length $250$, the agent is reset to a random position and a random goal is chosen.
Importantly, $3$ rectangular obstacles of random size are also created and are set to random positions.
Thus, the environment changes at the end of each episode.
A simple RRT planner (we use an implementation by \citet{zeng2019rrt}) then plans a feasible sequence of length $20$ through the state space.
Three random configurations of the environment are shown in \Figref{fig:moving_obstacles_env_0} to \Figref{fig:moving_obstacles_env_2}.

The resulting plan is input to the L2E agent, and is given in full to the plan-conditioned L2E policy.
We use the $S^\mathrm{bias}_{100,1000}$ strategy here.
We compare the L2E agent to using a HER agent with replay strategy ``Future 5''.
The HER policy is only given the goal as input.

\subsubsection{Results}
The results are shown in \Figref{fig:moving_obstacles_results}.
The HER agent learns a goal-conditioned policy, and does not have access to the plan.
Thus, it is unable to effectively generalize to a changed environment.
In contrast, the L2E agent learns to execute the entire plan.
It can generalize to unseen environments as long as it is provided with feasible plans by the RRT planner.
As a result, the L2E agent consistently shows success rates around 90\%.

\subsection{Optimizing HER and L2E replay strategy}
\label{sec:her_and_l2e_strategy}
For the basic pushing example (see \Figref{fig:pushing_setup}), we optimize the replay strategies of both HER and L2E to allow for a fair comparison.

\begin{figure}[h]
  \centering
  \subfloat[\label{fig:best_strategy_her}]{%
  \includegraphics[width=0.52\linewidth]{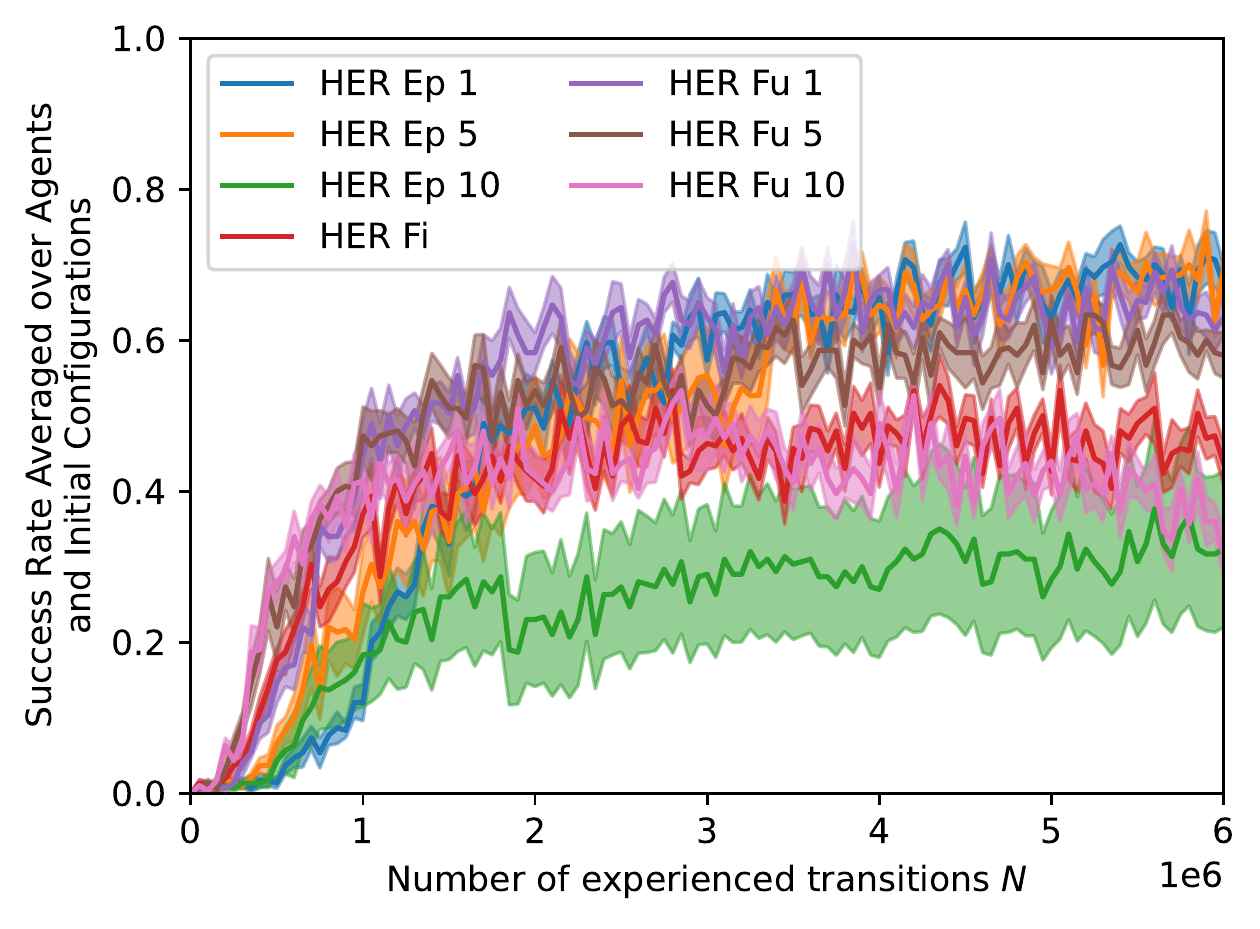}
  }
  \subfloat[\label{fig:best_strategy_l2e}]{%
          \includegraphics[width=0.485\linewidth]{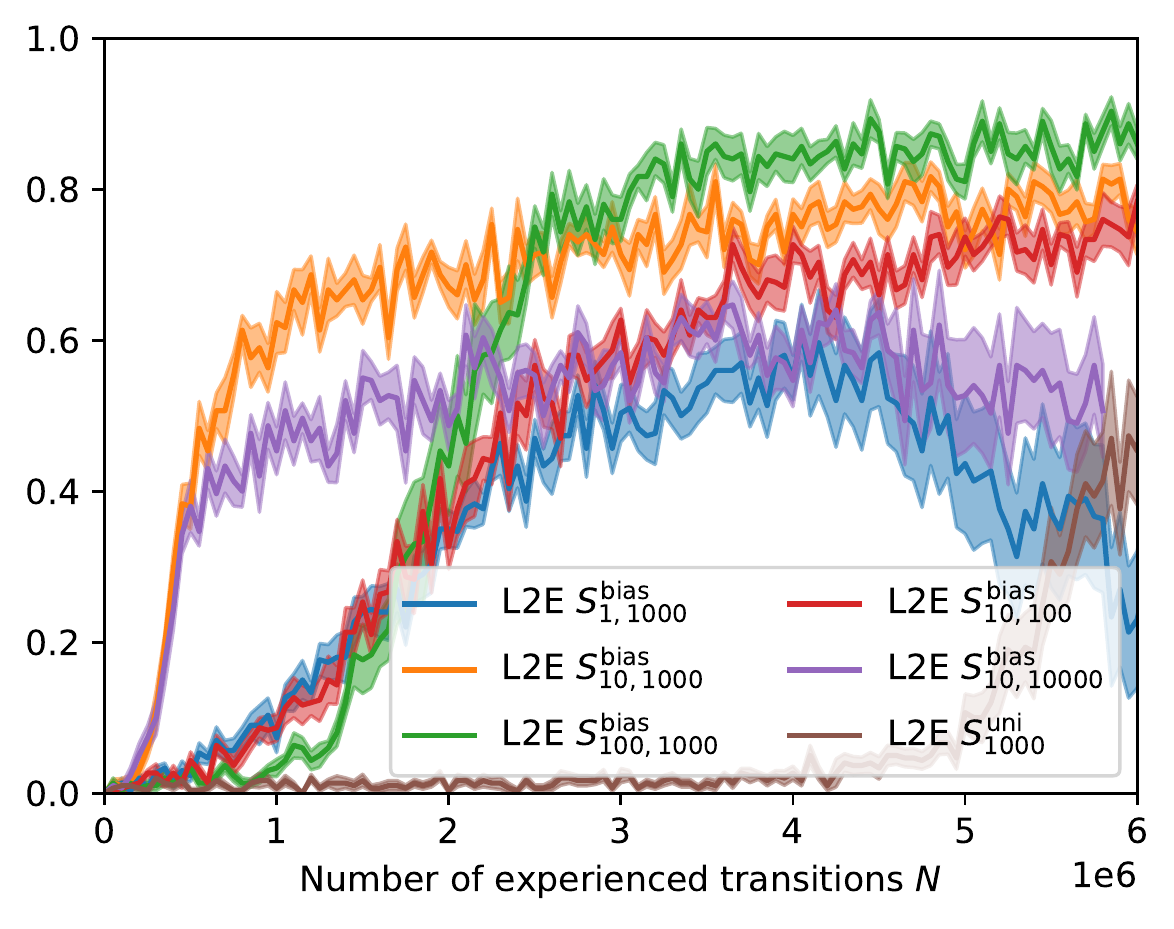}
  }
  \caption{
          For the basic pushing example (see \Figref{fig:pushing_setup}), we optimize the replay strategies of both HER and L2E to allow for a fair comparison.
          (a) For HER, we find that the ``Episode 1'', ``Episode 5'', and ``Future 1'' replay strategies result in comparable long-term performance, but ``Future 1'' seems to allow for slightly faster learning in the beginning.
          (b) For L2E, we find that using biased replay significantly improves sample efficiency.
          The L2E agents with uniform replay strategy $S^\mathrm{uni}_{1000}$ only catches up to the biased agents at $N \approx 8 \times 10^6$ (not shown).
          Out of the strategies tested, $S^\mathrm{bias}_{100,1000}$ performed best.
  }
  \label{fig:optimizing_strategies}
\end{figure}

\subsubsection{Optimizing HER Replay Strategy}
To allow for a fair comparison, we spent a considerable amount of effort to optimize the HER replay strategy.
Multiple experiments comparing different choices for the HER replay strategy are shown in \Figref{fig:best_strategy_her}.
Please also refer to \citet{andrychowicz2017} for a detailed description of the sampling strategies listed.
We find that using one future achieved goal from the same episode (``Fu 1'') for replay seems to result in the best long-term performance.
The strategies ``Ep 1'' and ``Ep 5'' perform comparably well.

\subsubsection{Optimizing L2E Replay Strategy}
We compare different replay strategies for L2E in \Figref{fig:best_strategy_l2e}.
We find that using biased sampling is crucial to achieve high sample efficiency.
The uniform strategy $S^\mathrm{uni}_{1000}$ catches up to the biased agent only after ca.\, 8 Million transitions (not shown in the figure).
However, overly pronounced biasing is also disadvantageous (see $S^\mathrm{bias}_{10, 10000}$ in comparison to $S^\mathrm{bias}_{10, 1000}$).
In this case, the L2E agent does not learn to generalize well to the test scenario.

\subsection{Experiments on Plan Encoding}
\label{sec:experiments_plan_encoding}
For the results reported in \secref{sec:results}, we encode plans analytically using box positions as described in \secref{sec:plan_encoding}
We will investigate in the following if (1) an encoding should be used at all and (2) whether it can be learned as well.

\begin{figure}[h]
  \centering
  \vspace{-0.45cm}
  \subfloat[\label{fig:vae_plan_reconstruction_0}]{%
  \includegraphics[width=0.33\linewidth]{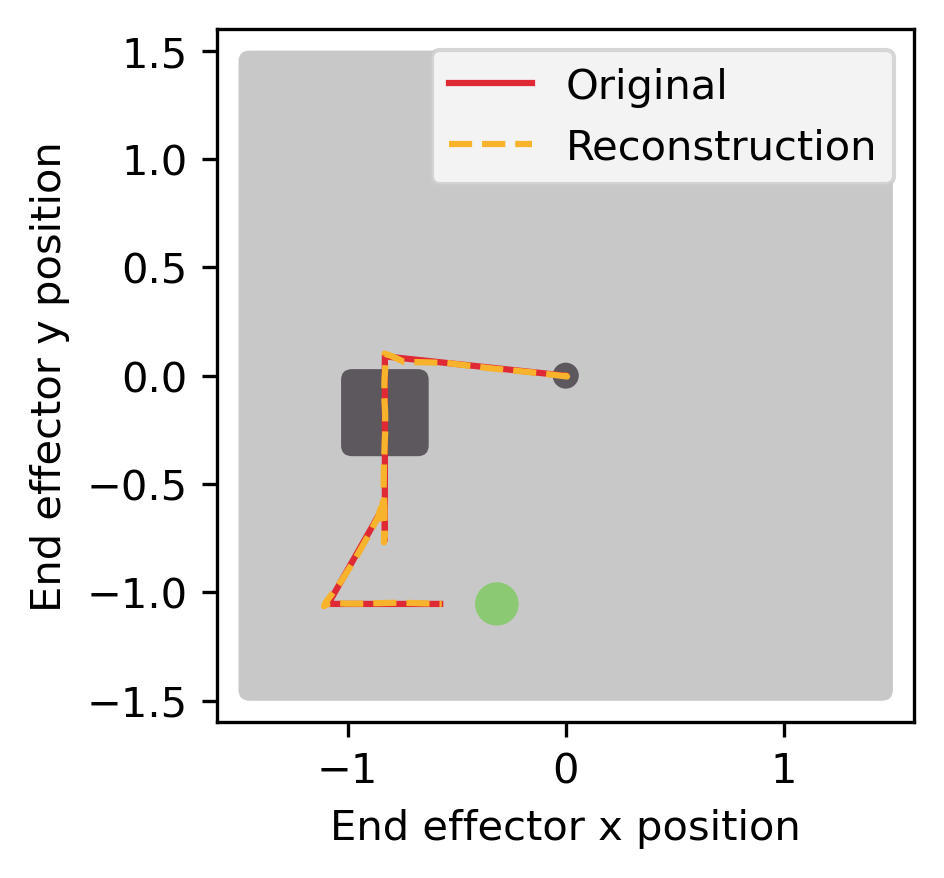}
  }
  \subfloat[\label{fig:vae_plan_reconstruction_1}]{%
          \includegraphics[width=0.33\linewidth]{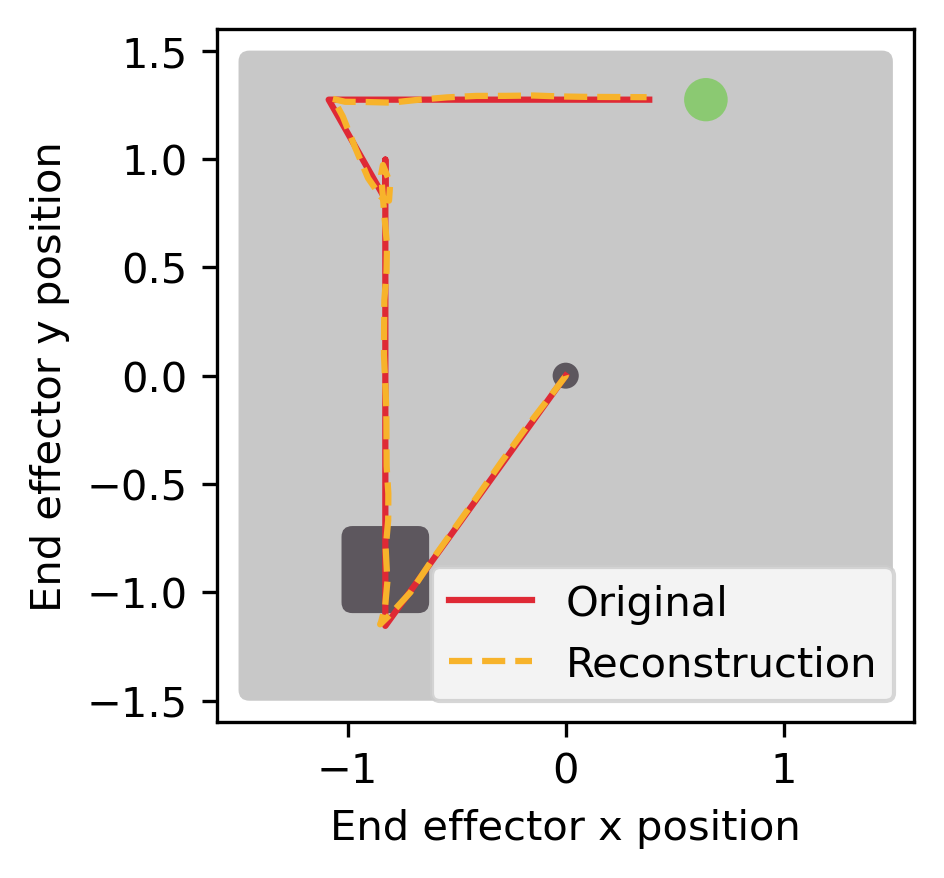}
  }
  \subfloat[\label{fig:vae_plan_reconstruction_2}]{%
          \includegraphics[width=0.33\linewidth]{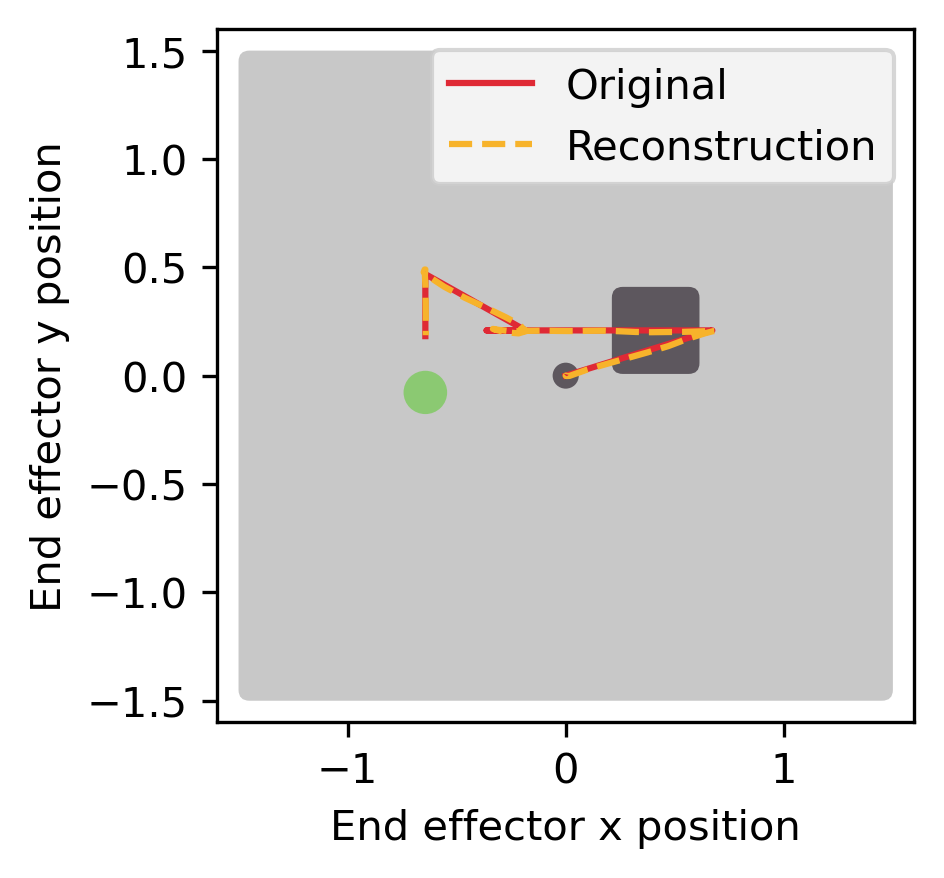}
  }
  \caption{
    Plans reconstructed by the trained VAE, together with the ground truth.
    The table is indicated in light gray, the initial configuration is indicated in dark gray, and the box's goal position is shown in green.
    The plans are sequences of length 50, containing 6D vectors of the planned positions of end effector and box.
    Shown here are only planned x- and y-positions of the end effector.
  }
  \label{fig:vae_plan_reconstructions}
\end{figure}

We contrast three different ways to encode the plan for the basic pushing environment (see \Figref{fig:pushing_setup}):
\begin{enumerate}
  \item Analytical encoding into a 4D latent space (entries correspond to intermediate box positions)
  \item Learned encoding using a variational autoencoder (VAE) trained with mean squared error (MSE) reconstruction loss, again using a 4D latent space.
  \item No encoding (plans are given in full to the policy).
\end{enumerate}
For option 2, we use an encoder made up of 6 fully connected layers with sizes decreasing geometrically from 1024 to 32.
For the decoder, we use the same architecture in opposite direction.
\Figref{fig:vae_plan_reconstructions} shows plans reconstructed by the trained VAE, together with the ground truth.

\begin{figure}[h]
  \centering
  \subfloat[\label{fig:encoding_comparison}]{%
  \includegraphics[width=0.52\linewidth]{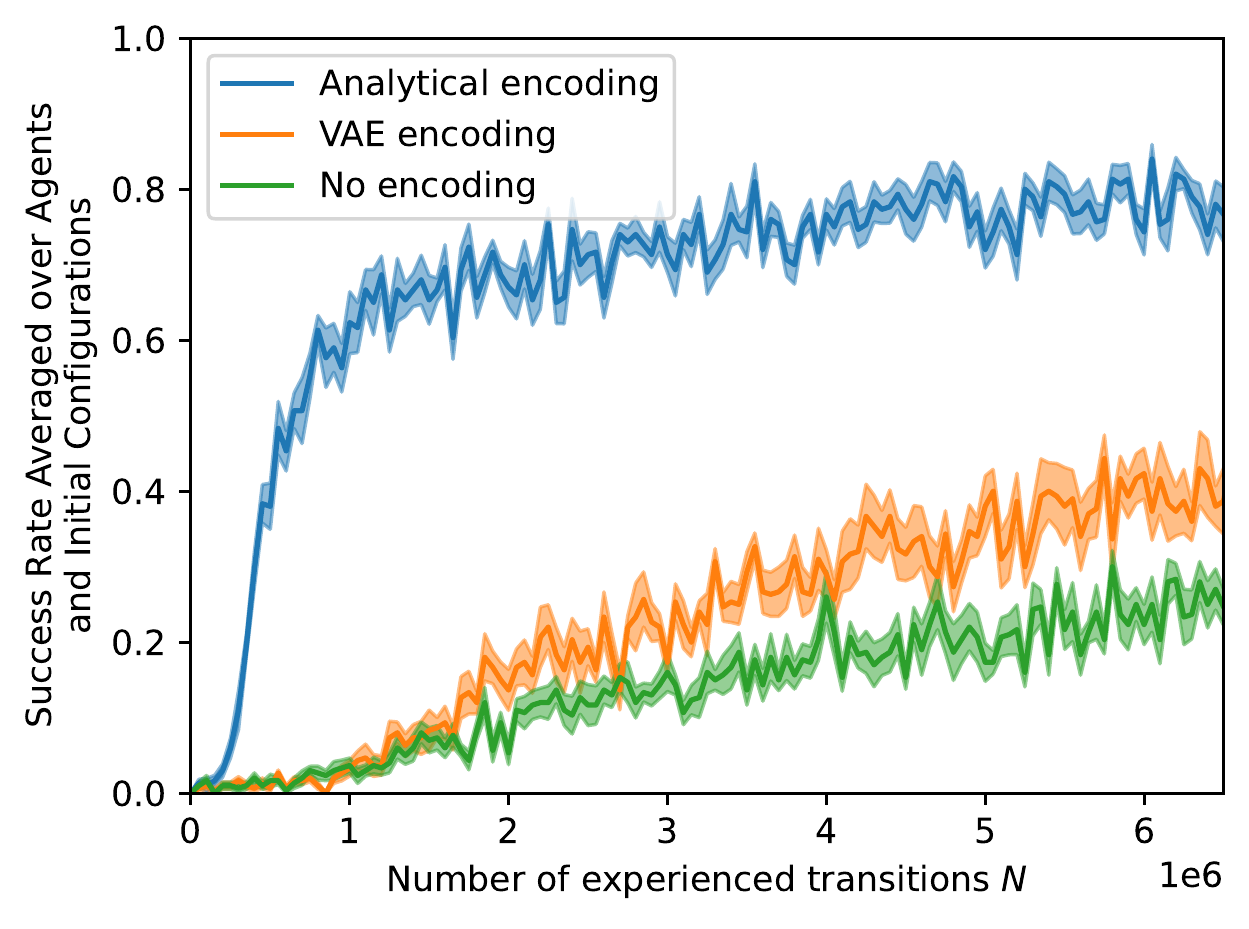}
  }
  \subfloat[\label{fig:plan_density_comparison}]{%
          \includegraphics[width=0.485\linewidth]{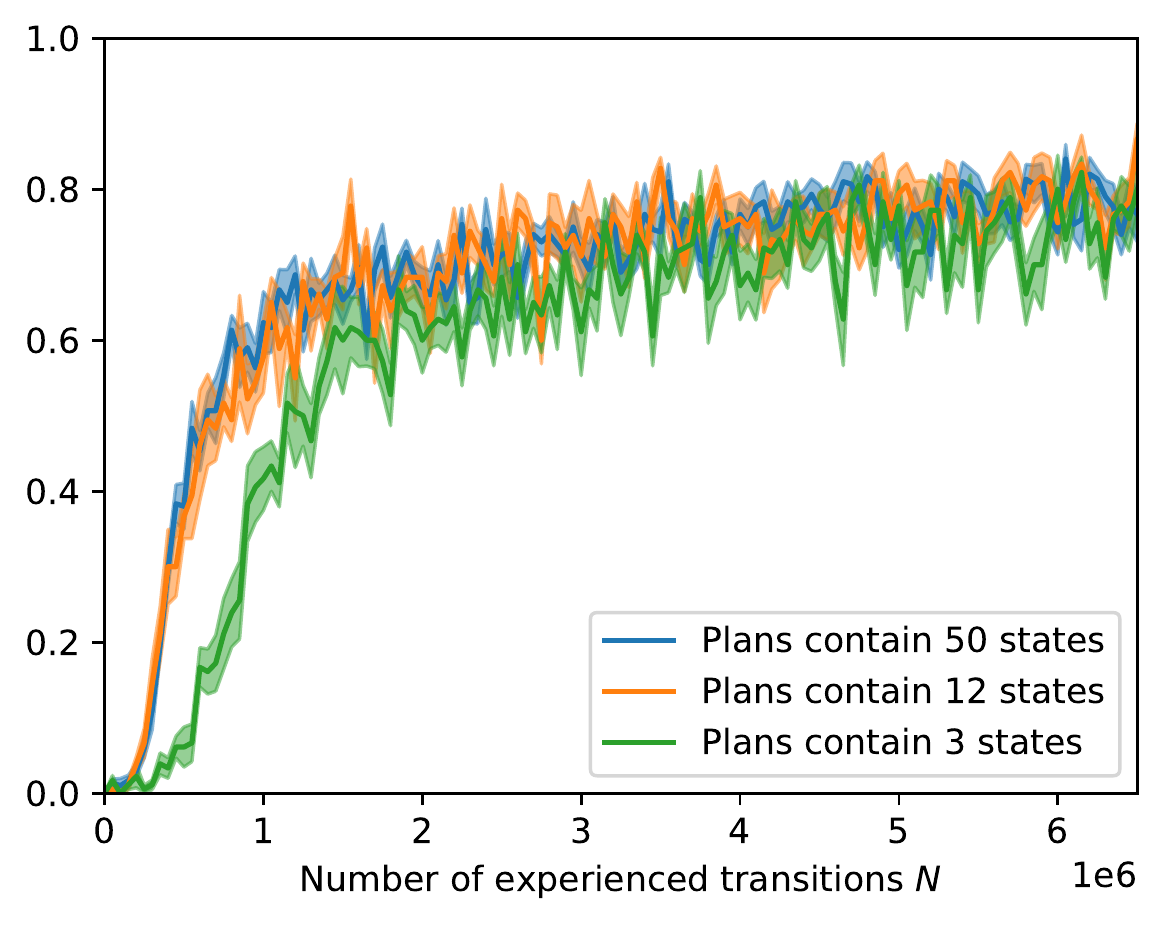}
  }
  \caption{Analysis of (a) plan encoding methods and (b) the influence of plan density for the L2E $S^\mathrm{bias}_{10, 1000}$ agent in the basic pushing environment (see \Figref{fig:pushing_setup}).
  (a) We find that using any encoding significantly outperforms using no encoding at all.
  Furthermore, using an analytical encoding, at least in this case, outperforms using a learned encoding of the same dimensionality.
  (b) We find that there is no significant difference between using 50 states and using 12 states.
  When using 3 states, the less informative reward signal leads to a slightly flatter learning curve in the beginning.
  Analytical encoding is used for all runs shown in (b).
  }
  \label{fig:ablation_encoding_and_density}
\end{figure}

For the basic pushing environment,
\Figref{fig:encoding_comparison} shows the results of the 3 options introduced above.

These results show that using an encoding is beneficial, even if it has to be learned. At least for the present example, using an analytical encoding was however more effective.

\subsection{Experiments on Plan Density}
\label{sec:experiments_plan_density}
Throughout this work, plans are represented as a trajectory of length $50$ for the basic pushing environment and $100$ for the obstacle pushing environment, consisting of 6D elements representing end effector position and box position.

Less dense plans result in a less informative reward shaping signal for the L2E agent.
To understand how much L2E relies on high-quality dense plans,
we compare training the L2E agent using planned state sequences of different density.
This comparison is performed in the basic pushing environment (see \Figref{fig:pushing_setup}).
For the same distribution of plans \Figref{fig:plan_density_comparison} shows the results for using $50$, $12$, or $3$ states as a representation.

We find that in this example, there is no significant difference between using $50$ states and using $12$ states.
When using $3$ states, the less informative reward signal leads to a slightly flatter learning curve in the beginning, but the agent seems to recover from this later on.

In this environment, L2E seems to be largely robust to lower plan densities.

\subsection{Basic Pushing Environment: Longer Runs for the best performing L2E Agent}
\label{sec:longer_run_l2e_100_1000_basic_pushing}
\begin{figure}[h]
  \centering
  \includegraphics[width=0.52\linewidth]{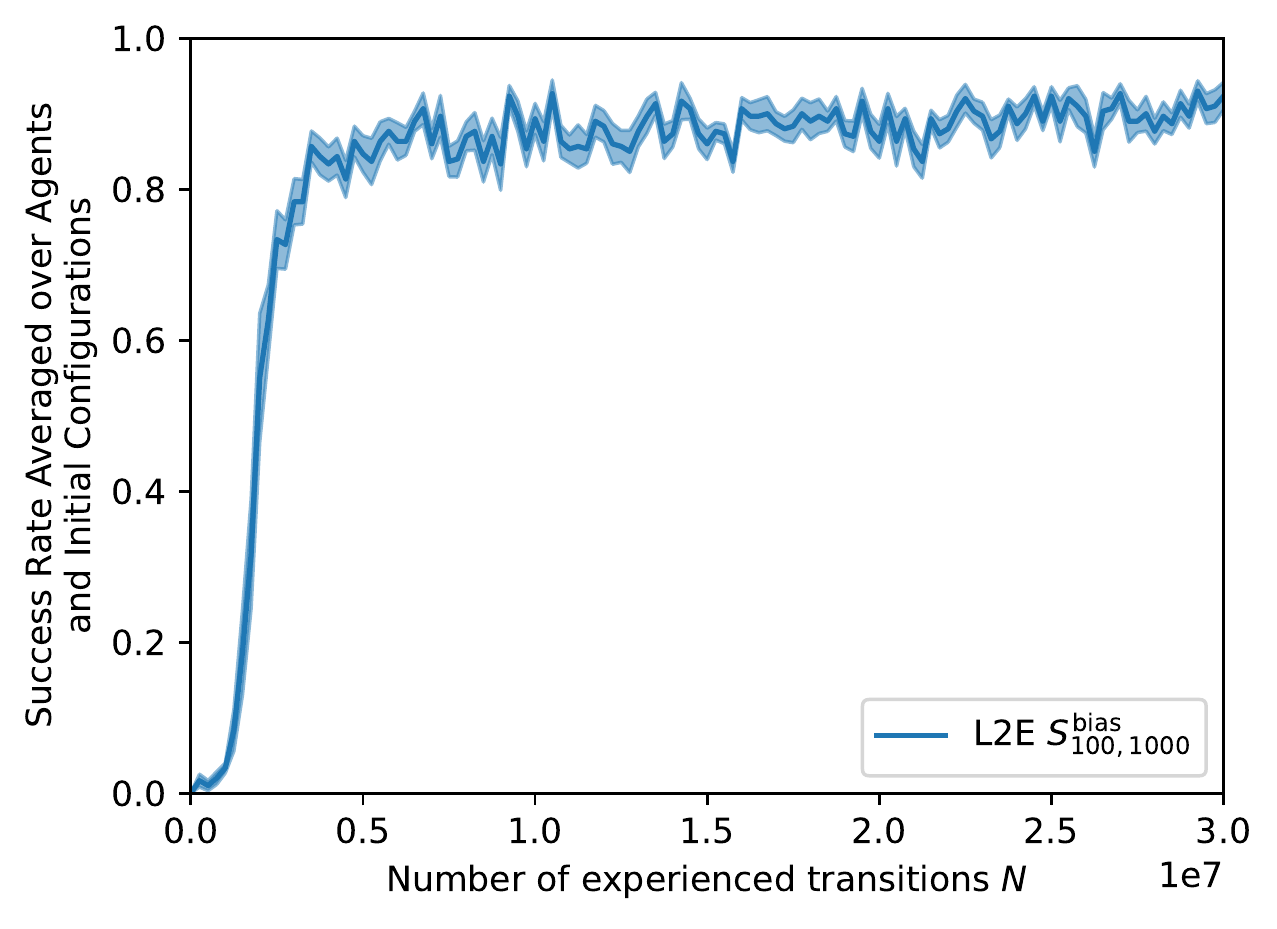}
  \caption{
  Results for running the L2E ${S^\mathrm{bias}}_{100, 1000}$ longer than shown in \Figref{fig:results_pushing}.
  The agent improves further.
  After $30$ million time steps, the mean success rate averaged over $10$ independently trained agents and $30$ randomized test rollouts is around $90\%$.}
  \label{fig:longer_run_l2e_100_1000_basic_pushing}
\end{figure}

\Figref{fig:longer_run_l2e_100_1000_basic_pushing} shows results for running the L2E ${S^\mathrm{bias}}_{100, 1000}$ longer than shown in \Figref{fig:results_pushing}.
The agent improves further.
After $30$ million time steps, the mean success rate averaged over $10$ independently trained agents and $30$ randomized test rollouts is around $90\%$.

\section*{Checklist}

\begin{enumerate}

\item For all authors...
\begin{enumerate}
  \item Do the main claims made in the abstract and introduction accurately reflect the paper's contributions and scope?
    \answerYes{}{Our claims made in the abstract and introduction are demonstrated in the experiments section.}
  \item Did you describe the limitations of your work?
    \answerYes{}{See \secref{sec:discussion}.}
  \item Did you discuss any potential negative societal impacts of your work?
    \answerNA{}{We see no immediate societal impact of our work.}
  \item Have you read the ethics review guidelines and ensured that your paper conforms to them?
    \answerYes{}{}
\end{enumerate}

\item If you are including theoretical results...
\begin{enumerate}
  \item Did you state the full set of assumptions of all theoretical results?
    \answerYes{}{We state the requirements for the background methods to be applicable in \secref{sec:background}, for L2E to be applicable in \secref{sec:method_learning_to_execute}, and further discuss them in \secref{sec:discussion}.}
	\item Did you include complete proofs of all theoretical results?
    \answerNA{}{We do not state formal theorems in this work.}
\end{enumerate}

\item If you ran experiments...
\begin{enumerate}
  \item Did you include the code, data, and instructions needed to reproduce the main experimental results (either in the supplemental material or as a URL)?
    \answerYes{}{We include the entire open-source code base to fully reproduce all figures in this paper in the supplementary material, and at \href{https://github.com/ischubert/l2e}{github.com/ischubert/l2e}. This includes README files to explain how it can be used. Although this should be sufficient to easily reproduce all results in this paper, we will also make the data files containing the results shown in this paper available upon request.}
  \item Did you specify all the training details (e.g., data splits, hyperparameters, how they were chosen)?
    \answerYes{}{See \secref{sec:method} and \secref{sec:results}.}
	\item Did you report error bars (e.g., with respect to the random seed after running experiments multiple times)?
    \answerYes{}{All figures show mean values as well confidence intervals (standard deviation of the mean).}
	\item Did you include the total amount of compute and the type of resources used (e.g., type of GPUs, internal cluster, or cloud provider)?
    \answerYes{}{See \secref{sec:method}.}
\end{enumerate}

\item If you are using existing assets (e.g., code, data, models) or curating/releasing new assets...
\begin{enumerate}
  \item If your work uses existing assets, did you cite the creators?
    \answerYes{}{We use open-source libraries that are cited where used.}
  \item Did you mention the license of the assets?
    \answerYes{}{We mention in the text that the assets are open source. More details on the exact license can be found in the references.}
  \item Did you include any new assets either in the supplemental material or as a URL?
    \answerYes{}{See question 3(a).}
  \item Did you discuss whether and how consent was obtained from people whose data you're using/curating?
    \answerNA{}{We do not use external data assets.}
  \item Did you discuss whether the data you are using/curating contains personally identifiable information or offensive content?
    \answerNA{}{We do not use external data assets.}
\end{enumerate}

\item If you used crowdsourcing or conducted research with human subjects...
\begin{enumerate}
  \item Did you include the full text of instructions given to participants and screenshots, if applicable?
    \answerNA{}{We do not use crowdsourcing or conducted research with human subjects.}
  \item Did you describe any potential participant risks, with links to Institutional Review Board (IRB) approvals, if applicable?
    \answerNA{}{We do not use crowdsourcing or conducted research with human subjects.}
  \item Did you include the estimated hourly wage paid to participants and the total amount spent on participant compensation?
    \answerNA{}{We do not use crowdsourcing or conducted research with human subjects.}
\end{enumerate}

\end{enumerate}

\end{document}